%% file: main.tex
\documentclass{article}
\usepackage{preprint,times}
\usepackage{authblk}
\usepackage{amsmath,amssymb,amsthm}
\usepackage{booktabs}
\usepackage{graphicx}
\usepackage{algorithm}
\usepackage{algpseudocode}
\usepackage{placeins}
\usepackage{microtype}
\usepackage{xcolor}
\usepackage{hyperref}
\hypersetup{hidelinks}
\usepackage{url}
\newcommand{\method}{\textsc{PACE}}
\newcommand{\mms}{\operatorname{MMS}}
\newcommand{\R}{\mathbb R}
\newtheorem{proposition}{Proposition}

\renewcommand{\and}{, }
\newcommand{\affiliations}[1]{\affil{#1}}

\title{PACE: Plug-and-Play Contextual Embedding for Feature Screening with Pretrained Tabular Foundation Models}

\author{%
Qi Qin,
Erbo Li,
Ting Wei,
Zizhou Huang,
Zixuan Qin,
Wu Wang,
Yifan Sun\thanks{Corresponding author.}%
}
\affiliations{Center for Applied Statistics, School of Statistics, Renmin University of China}

\begin{document}
\maketitle

\begin{abstract}
In high-dimensional tabular learning, feature screening provides a lightweight, model-agnostic way to remove irrelevant features before model fitting. However, scoring raw values directly can miss nonlinear or distributional structure. We introduce PACE (Plug-and-Play Contextual Embedding), which inserts a frozen tabular foundation model (TFM) column encoder before an existing feature-scoring rule, expanding each feature into a higher-dimensional contextual representation. Across controlled studies, PACE improves raw-space screening of complex nonlinear dependence with only modest additional encoding cost. {These gains translate into downstream predictive improvements on both TALENT and SelectArena datasets, where PACE generally outperforms matched raw-space screening across diverse downstream learners and feature budgets.} Matched random-weight and random-feature controls show that PACE gains from pretrained structure beyond generic dimensional expansion. PACE further achieves favorable performance--time trade-offs against task-fitted selectors and attribution-based methods, positioning pretrained column geometry as a reusable upstream primitive for tabular learning.
\end{abstract}

\section{Introduction}
\label{sec:intro}
High-dimensional tabular data are prevalent in scientific and industrial applications. Increasing dimensionality exacerbates estimation error, feature-acquisition costs, and computational burden~\citep{fan2011nonparametric}. \emph{Feature screening} addresses this problem by ranking columns (features) before model fitting and retaining a compact candidate set that can be reused across tree models, neural networks (NNs), and tabular foundation models (TFMs). Despite rapid progress in TFM scaling, high-dimensional tables remain a challenging regime~\citep{purucker2026beyond}. For a concrete reference point, TabPFN v2.5 supports datasets with up to 2,000 features~\citep{grinsztajn2025tabpfn}, while specialized approaches such as TabPFN-Wide~\citep{kolberg2025tabpfn} and GOTabPFN~\citep{habib2026gotabpfn} accommodate much wider inputs by modifying the predictive pipeline through continued pretraining or feature ordering and compression. Rather than modifying the predictor to accommodate increasingly wide inputs, we take an upstream approach: deciding which original columns should reach the predictive pipeline before a downstream predictor is chosen.

Classical marginal screening assigns each feature a \emph{utility} $U(X_j,Y)$ that quantifies its association with the response. The choice of $U$ determines which aspects of dependence are emphasized. SIS, for example, ranks features by marginal correlation~\citep{fan2008sure}; when $X$ is symmetric, the correlation between $X$ and $Y=X^2$ can vanish despite statistical dependence. More flexible utilities, such as distance correlation, capture broader forms of dependence~\citep{li2012feature}. However, the choice of utility is only part of the problem; real-world tables contain heterogeneous inputs: continuous and categorical features have different structures, and missing entries alter the empirical distribution seen by a utility. Nonlinear, category-aware, and missingness-aware screeners address important cases~\citep{cui2015model,li2024feature,bai2026model}, but each still operates on a particular representation. This suggests a \emph{representation bottleneck}: some screening failures may reflect the geometry presented to the utility rather than the utility alone. Pretrained TFMs offer a learned alternative to prespecified transformations. Broad pretraining yields transferable representations, and recent work has studied task-agnostic embeddings at multiple granularities, including row- and column-level representations~\citep{hoppe2025comparing,vogel2026towards}. We instead use a frozen per-column representation as input to an existing marginal screening utility. This leads to our central question: \emph{can pretrained column geometry improve classical screening across diverse data regimes without exposing the encoder to the current-task response?}

We introduce \method{}, a \textbf{P}lug-\textbf{A}nd-Play \textbf{C}ontextual \textbf{E}mbedding adapter for feature screening. \method{} inserts a frozen TFM column encoder before an existing feature-scoring rule, lifting each raw scalar feature into a higher-dimensional per-observation representation. \method{} leaves the rest of the screening pipeline unchanged while replacing raw-space geometry with pretrained column geometry. The resulting representation can expose nonlinear and distributional structure that is difficult to capture directly from raw values, improving feature ranking across heterogeneous and incomplete tabular data. \method{} is compatible with a broad range of screening utilities and downstream predictive models.

Our contributions are threefold. \textbf{First}, we introduce PACE as a plug-and-play representation adapter for existing feature-screening utilities, producing screened tables reusable across downstream predictors while preserving original feature identity. \textbf{Second}, extensive experiments across screening utilities and downstream predictors demonstrate broad empirical benefits. PACE improves over its raw counterpart in 98.6\% of classification and 83.9\% of regression non-tied utility--setting comparisons. {These gains translate into improved downstream prediction on nuisance-contaminated TALENT datasets and are further supported by results on the natural SelectArena benchmark, where PACE-DC and PACE-HSIC achieve statistically supported improvements over their raw counterparts.} \textbf{Third}, controlled representation ablations provide evidence that dimensional expansion alone does not explain these gains: pretrained representations improve over matched random-weight, projection, and random-feature controls, while backbone and extraction choices further affect screening performance.

\section{Related Work}
\label{sec:related}

\paragraph{Marginal and representation-enhanced screening.}
Feature screening is a classical approach to ultrahigh-dimensional data. Sure independence screening (SIS) uses marginal correlation~\citep{fan2008sure}, while DC extends screening through distance correlation~\citep{li2012feature}. Subsequent work adapts screening to heterogeneous feature types~\citep{cui2015model,xie2020category,li2024feature}, heavy-tailed designs~\citep{joudah2025air,jiang2025robust}, missing responses and censoring~\citep{bai2026model,zhang2026fdr}, and robust settings~\citep{guo2022stable,qin2025label}. Projection correlation (PC)~\citep{liu2022model} and improved projection correlation (IPC)~\citep{lin2026model} provide multivariate dependence utilities. A complementary direction transforms inputs before screening: kernel screening, random Fourier features (RFF), and InterDependence Scores (IDS) use prespecified kernel or feature maps~\citep{poignard2022feature,rahimi2007random,radhakrishnan2025efficiently}; DeepFS learns a dataset-specific autoencoder representation~\citep{li2023deep}; and large language models provide task-agnostic row embeddings for downstream tasks~\citep{hoppe2025comparing,vogel2026towards}. PACE instead provides a plug-and-play representation adapter for existing screening methods.

\paragraph{Task-adaptive feature selection.}
Feature selection typically couples dimensionality reduction with current-task fitting. LASSO uses an $\ell_1$ penalty~\citep{tibshirani1996regression}; Stochastic Gates~\citep{yamada2020feature}, LassoNet~\citep{JMLR:v22:20-848}, and Deep Lasso~\citep{cherepanova2023performance} learn differentiable or neural selectors. GradEnFS~\citep{liu2024supervised} aggregates input gradients, SAND~\citep{pad2025sand} uses additive-noise distortion, and xRFM-AGOP~\citep{beaglehole2026xrfm} uses gradient outer products. Unlike the encoder in PACE, these selectors tailor their model to the specific task.

\paragraph{Model-based feature attribution.}
Model-based feature scores include tree importance~\citep{breiman2001random}, SHAP~\citep{lundberg2017unified}, leave-one-covariate-out importance~\citep{lei2018distribution}, and Shapley Additive Global Importance~\citep{covert2020understanding}. Tabular transformers can expose marginal effects~\citep{thielmann2025beyond}, while ExplainerPFN meta-learns zero-shot Shapley-style importance without target-model queries~\citep{fonseca2026explainerpfn}. Attribution scores provide feature rankings for screening; rankings derived from a task-fitted predictor are specific to that predictor.

\paragraph{Learning on wide tables.}
Tabular foundation models such as TabPFN~\citep{grinsztajn2025tabpfn}, TabICL~\citep{qu2026tabiclv2}, and TabDPT~\citep{ma2025tabdpt} have expanded the supported feature count through pretraining. Approaches for wider tables include TabPFN sketching~\citep{feuer2023scaling}, continued pretraining~\citep{kolberg2025tabpfn}, feature ordering and compact tokenization~\citep{habib2026gotabpfn}, inference-time feature construction~\citep{nishikawa2025nonlinear}, and test-time partitioning~\citep{ye2026closer,team2026xiaomi}. More recent models~\citep{tao2026mitra,grinsztajn2026tabpfn,eo2026exaonetabular10} have adopted feature--screening mechanisms. PACE focuses on improving the representation used by screening utilities, with the resulting ranking reusable across downstream predictors.

\section{Methodology}
\label{sec:method}

\subsection{Setup}
Let $\mathcal D=\{(\boldsymbol x_i,y_i)\}_{i=1}^N$ contain $N$ observations (table rows) and $p$ candidate features (table columns), with a classification or regression response $Y=(y_1,\ldots,y_N)^\top$. Write $X_j=(x_{1j},\ldots,x_{Nj})^\top$ for column $j$, which may be continuous or categorical. The sparse target support is
\begin{equation}
\mathcal S=\left\{j:F_{Y\mid\boldsymbol X}(\cdot\mid\boldsymbol x)\text{ depends on }x_j\right\},
\qquad s=|\mathcal S|\ll p.
\label{eq:target}
\end{equation}
A conventional marginal screener assigns $\omega_j=\mathcal U(X_j,Y)$ and retains the top $k$ columns or thresholds the scores, with the cutoff determined by a cost budget or false discovery rate control~\citep{guo2023threshold, cai2026knockoff}. SIS uses absolute sample correlation; other utilities use different measures of dependence.

\subsection{Plug-and-Play Contextual Embedding}
Classical nonlinear screeners use a prespecified dependence geometry, which may not expose all nonlinear or distributional patterns equally well. \method{}~(Plug-and-Play Contextual Embedding) instead adapts the feature representation before applying a chosen screening utility. Specifically, a frozen column-contextual encoder maps each feature column to a higher-dimensional representation:
\begin{equation}
G_{\theta,N}:\mathcal X^N\rightarrow\R^{N\times d},
\qquad
\mathbf H_j=G_{\theta,N}(X_j),
\label{eq}
\end{equation}
{where $\mathcal X$ is the input space of a single feature} and $\mathbf H_j$ represents feature $j$ through $d$ latent coordinates; {screening assigns one score per feature and retains original columns for downstream prediction}. $G_{\theta,N}$ is column-local, permutation equivariant over observations, frozen across tasks, and never receives task responses. Categorical and missing values undergo the underlying TFM's native preprocessing.

PACE accommodates both vector- and scalar-input screening utilities: the former score $\mathbf H_j$ jointly, while the latter score each coordinate separately, with the resulting scores averaged:
\begin{equation}
\omega_j^{\method}
=
\begin{cases}
\mathcal U(\mathbf H_j,Y),
&  \text{vector-input utility (e.g., DC)},\\[1mm]
\dfrac{1}{d}\displaystyle\sum_{m=1}^{d}
\mathcal U\!\left(\mathbf H_j^{(:,m)},Y\right),
&\text{scalar-input utility (e.g., SIS)}.
\end{cases}
\label{eq:utility_dispatch}
\end{equation}

\textbf{Vector-input example: DC.} Distance correlation~\citep{li2012feature} uses pairwise Euclidean distances between the full $d$-dimensional row vectors of $\mathbf H_j$ to measure dependence on the response:
\begin{equation}
\widehat\omega_j^{\method\text{-DC}}
=
\widehat{\operatorname{dCor}}_N
\!\left(\mathbf H_j,\psi(Y)\right),
\label{eq:pace_dc}
\end{equation}
where $\psi(y)=y$ for regression, and $\psi(y)$ is a one-hot encoding of $y$ for classification. Using the doubly centered Euclidean-distance matrices $A_j$ and $B$ of $\mathbf H_j$ and $\psi(Y)$, we compute $\widehat{\operatorname{dCor}}_N=\{\langle A_j,B\rangle_F/(\|A_j\|_F\|B\|_F)\}^{1/2}$ (zero if the denominator vanishes), {where $\langle A,B\rangle_F=\sum_{i,\ell}A_{i\ell}B_{i\ell}$ and $\|A\|_F=\sqrt{\langle A,A\rangle_F}$ denote the Frobenius inner product and norm. }

\textbf{Scalar-input example: SIS.} For regression, SIS~\citep{fan2008sure} computes the absolute value of the sample Pearson correlation $\widehat\rho$ between each embedding coordinate and $Y$ across the $N$ observations, then averages the $d$ scores:
\begin{equation}
\widehat\omega_j^{\method\text{-SIS}}
=
\frac{1}{d}\sum_{m=1}^{d}
\left|\widehat\rho\!\left(\mathbf H_j^{(:,m)},Y\right)\right|.
\label{eq:pace_sis}
\end{equation}
For classification, each coordinate instead uses its largest absolute correlation with a one-versus-rest class indicator, followed by the same coordinate average. We also evaluate replacing this coordinate average with a maximum for PACE-SIS in Appendix~\ref{app:sis_aggregation}.

The original methods score raw features, whereas PACE applies the same utilities in a higher-dimensional embedding space. This parallels kernel methods, which map inputs into spaces where nonlinear structure can become more accessible~\citep{cristianini2000introduction, balasubramanian2013ultrahigh}. Rather than relying on a prespecified feature map, PACE uses a frozen column encoder pretrained on large collections of synthetically generated tabular tasks, allowing the representation to capture richer nonlinear and distributional regularities learned across diverse data-generating processes. The screening utility itself remains unchanged. Algorithm~\ref{alg:pace} and {Figure~\ref{fig:pace_workflow}} (Appendix~\ref{app:algorithm}) detail the procedure, while Appendix~\ref{app:utility_formulas} defines the utilities. The following proposition formalizes the information boundary of this response-blind mapping.

\begin{proposition}[No creation of marginal information]
\label{prop:no_creation}
For a fixed feature $j$, let $G_N$ be a fixed deterministic measurable column operator between standard Borel spaces.
If $X_j\perp Y$, then $G_N(X_j)\perp Y$.
If $G_N$ is injective on a set of $X_j$-probability one, then
$\sigma(G_N(X_j))=\sigma(X_j)$ up to null sets, where $\sigma(\cdot)$ denotes the generated sigma-algebra.
\end{proposition}

Proposition~\ref{prop:no_creation} shows that a fixed response-blind column operator preserves independence between the feature column and the response vector. Under almost-sure injectivity, the transformation also preserves the information in the original column while reorganizing its geometry. This provides a simple information-theoretic interpretation of PACE: gains can arise from making existing marginal structure more accessible to the screening utility.

\subsection{Column encoders and computational cost}
\paragraph{Extraction boundary and instantiation.}
PACE can be instantiated at column-aligned representation stages across different backbones and extraction depths. TabICL~\citep{qu2025tabicl,qu2026tabiclv2} and TabPFN v3~\citep{grinsztajn2026tabpfn} provide compatible column-aligned representations. The main experiments use a frozen TabICL v1 encoder, whose shared column-wise module processes each feature independently; cross-feature interaction occurs only in the subsequent row-wise module. Each column is projected to $d=128$ and processed by three induced self-attention blocks with $M=128$ inducing points, with the main representation read out cumulatively after the second block ($L_2$). Ablations vary backbone, depth, execution path, and pretrained versus random weights.

\paragraph{Computational cost.}
TabICL v1 column encoding costs $O(NMp)$, linear in $p$. SIS-type scoring costs $O(Npd)$; distance and kernel utilities cost $O(N^2pd)$ and can dominate encoding. Appendix~\ref{app:repro} specifies the implementation.

\section{Experiments}
\label{sec:experiments}
We evaluate PACE from three perspectives. First, matched raw-to-PACE comparisons isolate the representation effect by holding the screening utility and downstream learner fixed, with controlled simulations measuring support recovery and nuisance-contaminated TALENT datasets~\citep{JMLR:v26:25-0512} assessing downstream transfer. Second, natural high-dimensional datasets provide a complementary real-data evaluation. Third, representation ablations test whether the gains arise from pretrained geometry rather than dimensional expansion alone, while cross-paradigm comparisons place PACE alongside reusable and task-fitted selectors with different fitting and response-access requirements.

\subsection{Comparison methods and evaluation criteria}
\label{sec:evaluation_protocol}

\paragraph{Comparison methods.}
We organize the evaluated selectors by the mechanism that produces their ranking. Appendix~\ref{app:cross_paradigm} gives exact constructions, fitting protocols, and comparison scope. \textbf{Featurewise screening utilities.} The main text highlights matched raw/PACE versions of SIS, DC, PC, and HSIC. Appendix~\ref{app:raw_pace_simulation} lists the full set of utilities. We also include IDS~\citep{radhakrishnan2025efficiently} and other prespecified nonlinear mappings in Appendix~\ref{app:cross_paradigm}. \textbf{Fitted-model importance.} We rank features using importance scores from random forests~\citep{breiman2001random}, LightGBM~\citep{ke2017lightgbm}, and CatBoost~\citep{prokhorenkova2018catboost}. \textbf{Predictive attribution.} We compute SHAP values~\citep{lundberg2017unified} for LightGBM, CatBoost, and TabICL v2; TabICL-SHAP uses two estimators. We also evaluate Leave-One-Covariate-Out (LOCO)~\citep{lei2018distribution} and Shapley Additive Global Importance (SAGE)~\citep{covert2020understanding}. \textbf{Supervised learned selectors.} LassoNet~\citep{JMLR:v22:20-848}, Deep Lasso~\citep{cherepanova2023performance}, GradEnFS~\citep{liu2024supervised}, SAND~\citep{pad2025sand}, and xRFM-AGOP~\citep{beaglehole2026xrfm} learn task-specific rankings from the current labels.

\paragraph{Evaluation metrics.}
The TALENT study uses AUC for binary classification, macro-averaged AUC for multiclass classification, and RMSE for regression. Natural high-dimensional classification is evaluated using accuracy. Controlled simulation studies use the minimum model size (MMS),
\begin{equation}
\mms=\max_{j\in\mathcal S}r_j,
\label{eq:mms}
\end{equation}
where $r_j$ is the rank of feature $j$. MMS is the size of the smallest top-ranked set containing every active feature. Its oracle value is $s$, and lower is better. Table~\ref{tab:controlled_core} reports medians across matched replications; Appendix~\ref{app:controlled} additionally reports interquartile ranges (IQRs).

For {nuisance-contaminated} dataset $q$, let {$\mathcal I_q^{\mathrm{orig}}$} index its original features and let $\widehat{\mathcal S}_{k,q}$ be the retained set.  Original-feature retention is {$\operatorname{Ret}_q=|\widehat{\mathcal S}_{k,q}\cap \mathcal I_q^{\mathrm{orig}}|/|\mathcal I_q^{\mathrm{orig}}|$}; higher values mean that fewer injected nuisance coordinates displace original features. For learner $\ell$, repeated split $r$, and TALENT task metric $Q$, the matched PACE-minus-raw effect is
\begin{equation}
\delta_{\ell q}=\frac{1}{|\mathcal R_{\ell q}|}\sum_{r\in\mathcal R_{\ell q}}
\begin{cases}
Q^{\rm PACE}_{r\ell q}-Q^{\rm raw}_{r\ell q},&\text{classification},\\
(\mathrm{RMSE}^{\rm raw}_{r\ell q}-\mathrm{RMSE}^{\rm PACE}_{r\ell q})/
\mathrm{RMSE}^{\rm clean}_{r\ell q},&\text{regression}.
\end{cases}
\label{eq:real_effect}
\end{equation}
Here, $\mathrm{RMSE}^{\rm clean}_{r\ell q}$ is the RMSE of learner $\ell$ on the corresponding clean, {non-nuisance-contaminated} table. We average $\delta_{\ell q}$ over the available matched learners within each dataset, then report the mean across datasets for classification and the median for regression. Uncertainty intervals are obtained by bootstrapping base datasets. The broader comparison reports improvability, $100(E_m-E_*)/E_m$, within dataset--learner blocks, where $E_m$ is the method's error and $E_*$ is the lowest error among the included methods; lower is better. Coverage and aggregation details appear in Appendix~\ref{app:real_full}.

\subsection{Controlled simulations across utilities}
\label{sec:mechanism}
The controlled simulation study uses 16 classification and nine regression settings with known active sets. It covers additive and nonlinear signals, imbalance, mixed feature types, correlated features with values missing completely at random (MCAR), and widths up to $100{,}000$. Regular settings use 50 matched replications; Appendix~\ref{app:controlled} gives the generators, counts, and complete results. 

\begin{table}[t]
\centering
\caption{\textbf{Selected controlled MMS: raw versus PACE-$L_2$.} Lower is better; $s$ is the oracle MMS and IDS is a standalone expanded baseline. The tables in the appendix list the results in more detail.}
\label{tab:controlled_core}
\begin{tabular}{lrrrrrr}
\toprule
Scenario & $s$ & SIS & DC & PC & HSIC & IDS\\
\midrule
\multicolumn{7}{l}{\textit{Classification}}\\
Additive Gaussian & 8 & $8\to8$ & $8\to8$ & $8\to8$ & $8\to8$ & 8\\
Nonlinear, independent & 4 & $2179\to4$ & $6\to4$ & $18\to7$ & $4\to5$ & 4\\
Severe imbalance & 10 & $1728\to10$ & $86\to21$ & $278\to72$ & $26\to18$ & 10\\
$p=100{,}000$ & 15 & $86096\to15$ & $15\to15$ & $15\to15$ & $15\to15$ & 15\\
Mixed, $K=25$ & 6 & $753\to62$ & $30\to8$ & $75\to8$ & $118\to20$ & 70\\
MCAR $0.30$ & 4 & $2128\to22$ & $20\to20$ & $130\to86$ & $7\to18$ & 5\\
\addlinespace
\multicolumn{7}{l}{\textit{Regression}}\\
\midrule
Nonlinear, Gaussian & 4 & $1461\to86$ & $54\to14$ & $92\to28$ & $49\to27$ & 1054\\
$p=100{,}000$ & 8 & $88153\to19$ & $164\to23$ & $2531\to97$ & $101\to53$ & 86604\\
Mixed, $K=5$ & 4 & $197\to6$ & $40\to12$ & $108\to108$ & $74\to58$ & 1556\\
Block, MCAR $0.15$ & 3 & $18\to11$ & $14\to12$ & $24\to58$ & $28\to54$ & 1064\\
\bottomrule
\end{tabular}
\end{table}

Table~\ref{tab:controlled_core} separates settings where raw screening is already sufficient from those where representation becomes limiting. In simple additive designs, nearly all raw utilities already attain the oracle MMS, and PACE retains this ability to recover simple signals. Under nonlinear or distributional signals, however, raw SIS can rank active features extremely late, whereas PACE-SIS often reduces MMS below 100 and frequently to the oracle level. PACE-SIS reduces MMS from 2,179 to 4 in nonlinear classification and, at $p=100{,}000$, from 86,096 to 15 in classification and from 88,153 to 19 in regression. More flexible utilities such as DC, HSIC, and PC also improve across diverse challenging settings, with utility-dependent gains.

IDS provides a useful representation-based comparator. Both methods enrich the feature representation before measuring dependence, but IDS uses a fixed finite-dimensional map whereas PACE uses a pretrained column Transformer. PACE's lower MMS in several difficult settings is consistent with the ablations, which indicate benefits from pretrained structure beyond generic dimensional expansion.

Beyond the representative utilities highlighted above, we aggregate results over the full utility--setting grid. In classification, 32.2\% of comparisons are unchanged, largely due to oracle-level ties in easier settings; among the remaining comparisons, 98.6\% favor PACE. In regression, only 1.6\% are unchanged, and PACE improves 83.9\% of the remainder.

\subsection{Nuisance-contaminated TALENT tables}
\label{sec:real}
We {contaminate} 288 TALENT datasets~\citep{JMLR:v26:25-0512} with independent $\mathcal N(0,1)$ nuisance columns\footnote{Numerical inputs are normalized before downstream prediction, so standard Gaussian noise provides a common, scale-controlled contamination stress test across datasets.} until the total width reaches 2,000. Screening retains $k=\min(N-1,3p_{\rm orig})$, where $N$ is the training-set size and $p_{\rm orig}$ is the number of original features before {contamination}. Multiclass datasets with more than 10 classes are excluded. The common budget compares ranking quality at equal retained dimensionality, while the $3p_{\rm orig}$ target allows slack beyond the original width. This construction provides an identifiable numerical-contamination stress test: original-feature retention directly measures how well a screener prevents injected nuisance variables from displacing the original columns. Appendix~\ref{app:real_protocol} documents preprocessing and sampling details. We evaluate the selected columns using ten learners in three families: \textbf{Tree models:} LightGBM~\citep{ke2017lightgbm}, XGBoost~\citep{chen2016xgboost}, and CatBoost~\citep{prokhorenkova2018catboost}; \textbf{NN models:} a standard MLP, RealMLP~\citep{holzmuller2024better}, and TabM~\citep{gorishniy2025tabm}; and \textbf{TFM models:} TabICL v2~\citep{qu2026tabiclv2}, TabPFN v3~\citep{grinsztajn2026tabpfn}, TabPFN v2.6~\citep{hollmann2025accurate}, and TabDPT~\citep{ma2025tabdpt}.

\subsubsection{Matched screening and downstream effects across utilities and learners}

\begin{figure}[t]
\centering
\includegraphics[width=.85\linewidth]{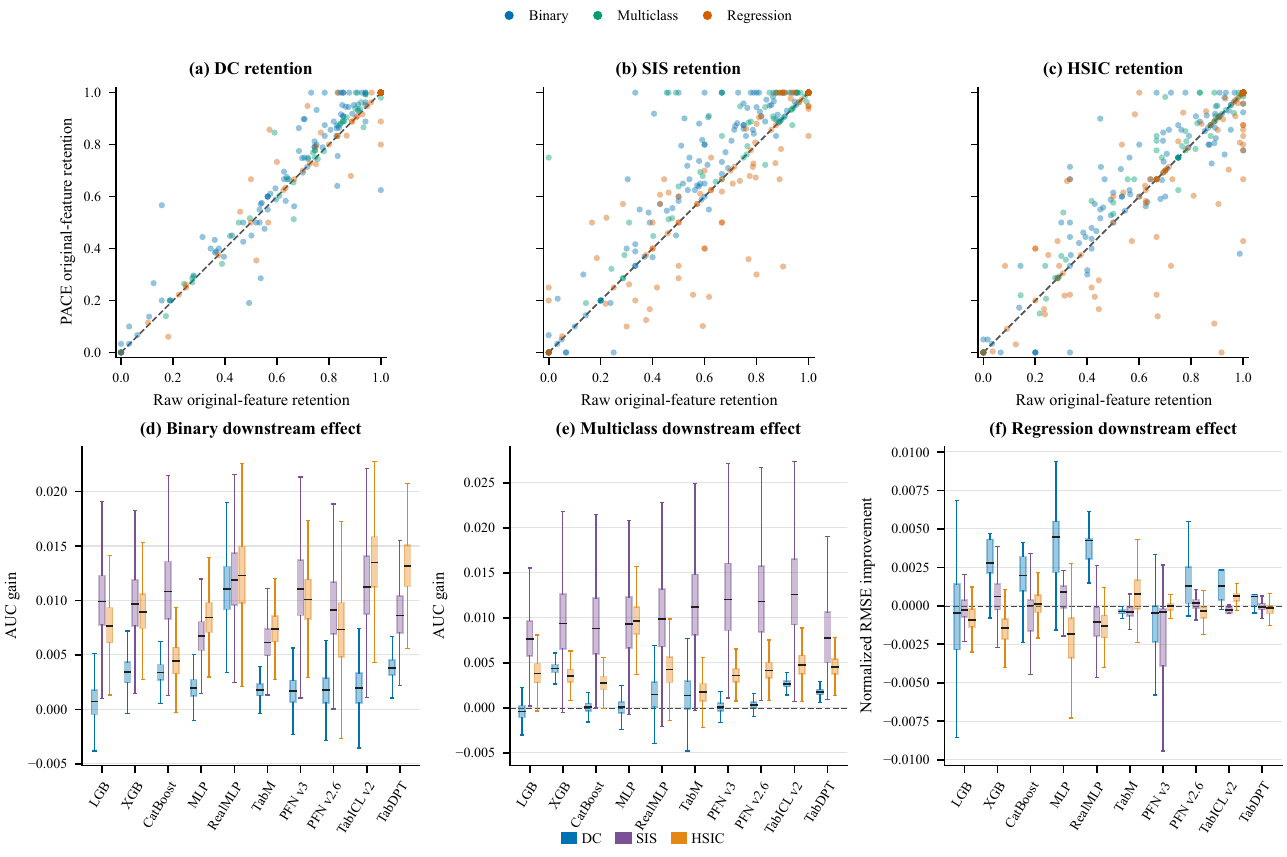}
\caption{\textbf{Nuisance rejection and downstream recovery on TALENT.} Top: matched original-feature retention; points above the diagonal favor PACE. Bottom: bootstrap distributions of learner-specific PACE-minus-raw effects; positive values favor PACE. Effects are binary/multiclass AUC gains or clean-normalized RMSE improvement.}
\label{fig:real_primary}
\end{figure}

The top row of Figure~\ref{fig:real_primary} quantifies nuisance rejection. Averaged over all datasets, PACE-SIS substantially increases original-feature retention from $0.671$ to $0.719$, while PACE-DC and PACE-HSIC yield more modest changes. These results highlight the complementary benefits of PACE across different screening utilities. The bottom row shows learner-specific downstream effects. Despite the already strong performance of raw DC and HSIC, their PACE variants provide further predictive improvements. {Averaged across ten downstream learners, PACE-SIS improves binary AUC by $0.010$ and multiclass macro-AUC by $0.011$, while PACE-DC achieves gains of $0.003$ and $0.001$, and PACE-HSIC achieves gains of $0.009$ and $0.005$, respectively (Tables~\ref{tab:talent_learner_binary} and \ref{tab:talent_learner_multiclass}).} In regression, improvements are generally more modest, with PACE-DC achieving a median normalized RMSE improvement of $0.002$, while SIS and HSIC exhibit smaller positive gains.

Appendix~\ref{app:talent_matched} further examines nuisance sensitivity across learner families. NNs and TFMs are substantially more affected than tree models: in binary classification, nuisance damage is $0.049$ for trees versus $0.124$ and $0.115$ for NNs and TFMs, while clean-normalized regression damage rises from $0.118$ for trees to $0.401$ and $0.633$, respectively. PACE reduces this burden before downstream prediction, with benefits extending across screening utilities and learner families. A complementary recovery analysis in Appendix~\ref{app:talent_matched} shows that PACE restores a larger share of the predictive performance lost to nuisance features.

\subsubsection{Comparison with classical and learned selectors}

\begin{figure}[t]
\centering
\includegraphics[width=.9\linewidth]{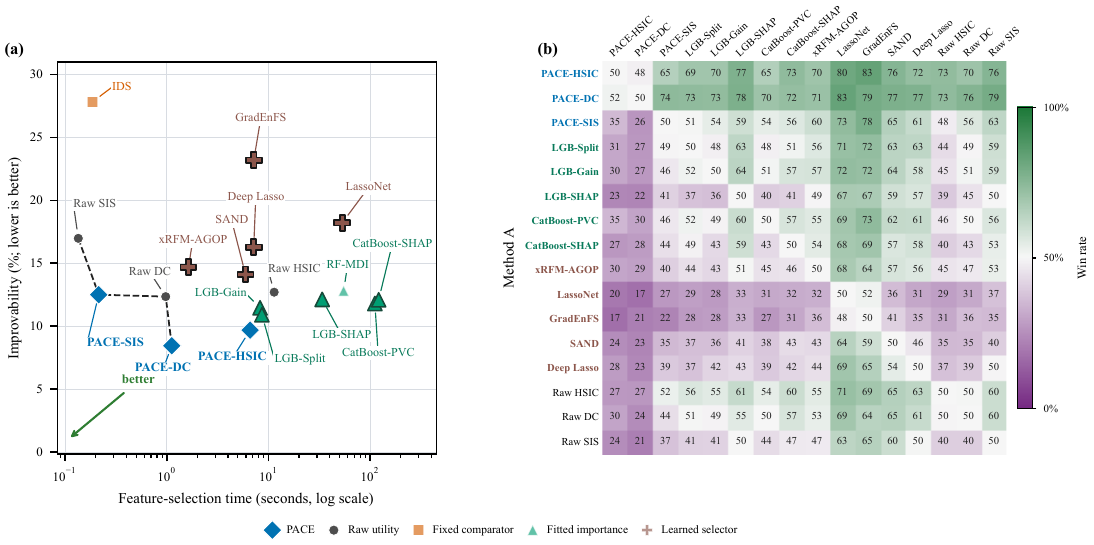}
\caption{\textbf{TALENT performance--time trade-offs and pairwise comparisons.} (a) Time vs. improvability (lower-left is better); the Pareto frontier is shown, and marker shapes indicate method families. (b) Win rate of the row method against the column method (ties count as half a win). Results are averaged over learners; protocols are given in Appendix~\ref{app:talent_field}.}
\label{fig:real_rank}
\end{figure}

For the broader comparison, we impose a three-hour screening-time budget and rank methods that complete within this limit. Unless otherwise noted, all feature-selection times are measured on a single NVIDIA A100 GPU. Figure~\ref{fig:real_rank}(a) shows that {PACE-SIS and PACE-DC} lie on the displayed performance--time Pareto frontier, spanning complementary operating points: PACE-SIS, PACE-DC, and PACE-HSIC achieve $12.5\%$, $8.5\%$, and $9.7\%$ improvability at screening times of $0.21$, $1.12$, and $6.60$ seconds, respectively. PACE-DC achieves both lower improvability and lower screening time than the evaluated learned selectors, including xRFM-AGOP and LassoNet, while PACE-DC and PACE-HSIC further improve the trade-off relative to tree-based feature-importance methods such as LightGBM gain and CatBoost PVC. Among methods completing within the budget, {PACE-DC} attains the lowest improvability and {ranks first in multiclass classification and regression, while PACE-HSIC ranks first in binary classification}. In Figure~\ref{fig:real_rank}(b), PACE-HSIC's aggregated win rates against the other displayed methods range from approximately $48\%$ to $83\%$. 

We next ask whether upstream screening also improves prediction with wide-table TFMs. Across the four general-purpose TFMs in Figure~\ref{fig:runtime_tfm_comparison}(a)--(b), screening increases mean classification AUC and reduces median relative RMSE gaps compared with the matched all-feature pipelines. On the shared classification support, PACE-HSIC with TabICL v2 achieves a task-balanced mean AUC of $0.891$, compared with $0.879$ for TabPFN-Wide and $0.753$ for GOTabPFN. For regression, using clean TabICL v2 as the reference, PACE-DC reduces the median relative RMSE gap from $17.82\%$ without screening to $1.31\%$; GOTabPFN yields a $101.85\%$ gap, while TabPFN-Wide does not provide regression support. These results show that upstream screening can complement general-purpose TFMs while remaining reusable across downstream models.

The computational contrast is even larger for more expensive attribution procedures. TabICL-SHAP, LOCO, and SAGE exceed the three-hour budget in this comparison, whereas every PACE variant completes screening in under ten seconds. Thus, PACE improves the performance--time trade-off without fitting a task-specific selector or repeatedly querying a predictive model. This large gap motivates the next experiment, where we examine how the screening cost of PACE and attribution-based alternatives scales as feature dimensionality increases.

\subsubsection{Feature-dimension scaling}
\label{sec:feature_scaling_wide_prediction}

\begin{figure}[t]
\centering
\includegraphics[width=.9\linewidth]{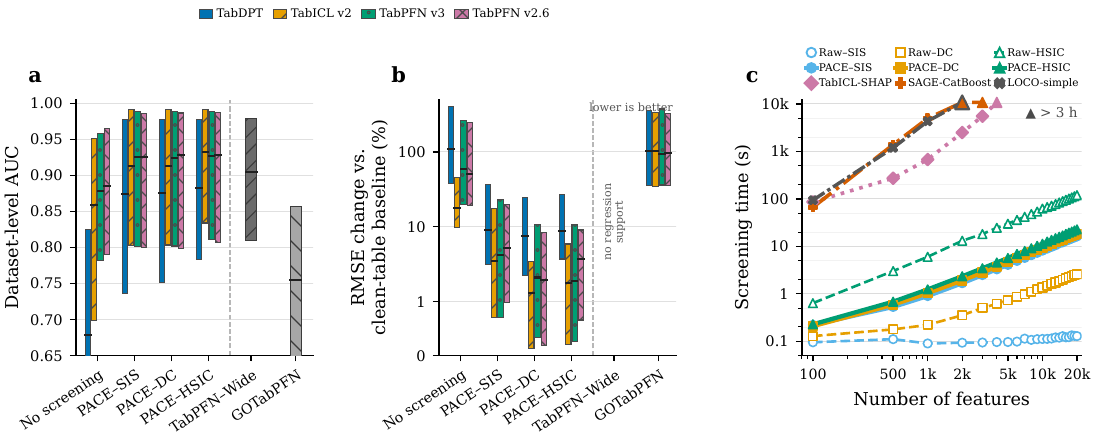}
\caption{\textbf{Runtime scaling and wide-table TFM prediction.} (a) Dataset-level AUC on classification datasets shared by all compared pipelines. (b) Relative RMSE change from the corresponding clean-table TFM on the regression datasets; lower is better. Boxes show quartiles and medians. (c) Screening time at $N=1{,}000$; triangles mark values above three hours.}
\label{fig:runtime_tfm_comparison}
\end{figure}

To examine how this computational gap evolves with dimensionality, we fix $N=1{,}000$ and vary the number of features $p$ from 100 to 20,000  using a consumer-grade NVIDIA RTX 4080 SUPER GPU (Figure~\ref{fig:runtime_tfm_comparison}(c)). The PACE variants and traditional screening utilities scale approximately linearly with $p$. In contrast, TabICL-SHAP and SAGE-CatBoost show roughly quadratic growth beyond their initial overhead, and LOCO-simple also scales steeply because it repeatedly refits a random forest after removing each feature. Consequently, the runtime gap widens rapidly as dimensionality increases: at $p=2{,}000$, the PACE variants require about 2 seconds, compared with $2{,}497$ seconds for TabICL-SHAP, $17{,}769$ seconds for SAGE-CatBoost, and $14{,}840$ seconds for LOCO-simple.  These results show that PACE retains a favorable screening-time profile as feature dimensionality grows, complementing the performance--time trade-offs observed in Figure~\ref{fig:real_rank}. Appendix~\ref{app:sample_runtime_scaling} additionally reports screening time as the sample size increases at fixed feature dimension.

\paragraph{Natural high-dimensional evaluation.}
We evaluate ten natural classification datasets with $500$--$10{,}304$ features at a 30\% selection budget, where TFMs can be sensitive to feature width~\citep{purucker2026beyond}. PACE-SIS with TabPFN v3 ($92.3\%$) and PACE-DC with TabICL v2 ($92.5\%$) have the highest complete-coverage means for their respective downstream models (Appendix Table~\ref{tab:highdim_accuracy}).

\subsection{{Predictive performance on SelectArena}}
\label{sec:selectarena}
{We evaluate PACE and raw SIS, DC, and HSIC against all 15 published SelectArena selectors on 71 natural real-world datasets~\citep{schafer2026selectarena}. Following the SelectArena protocol, we consider five feature budgets for each dataset and evaluate the selected subsets using four downstream predictors: TabICL v2, random forest (RF), LightGBM (LGBM), and a linear model. All methods are evaluated using three matched outer folds. We report error reduction relative to random selection and pairwise Elo ratings; further details of the experimental protocol are provided in Appendix~\ref{app:selectarena_protocol}.}

\begin{figure}[t]
\centering
\includegraphics[width=\linewidth]{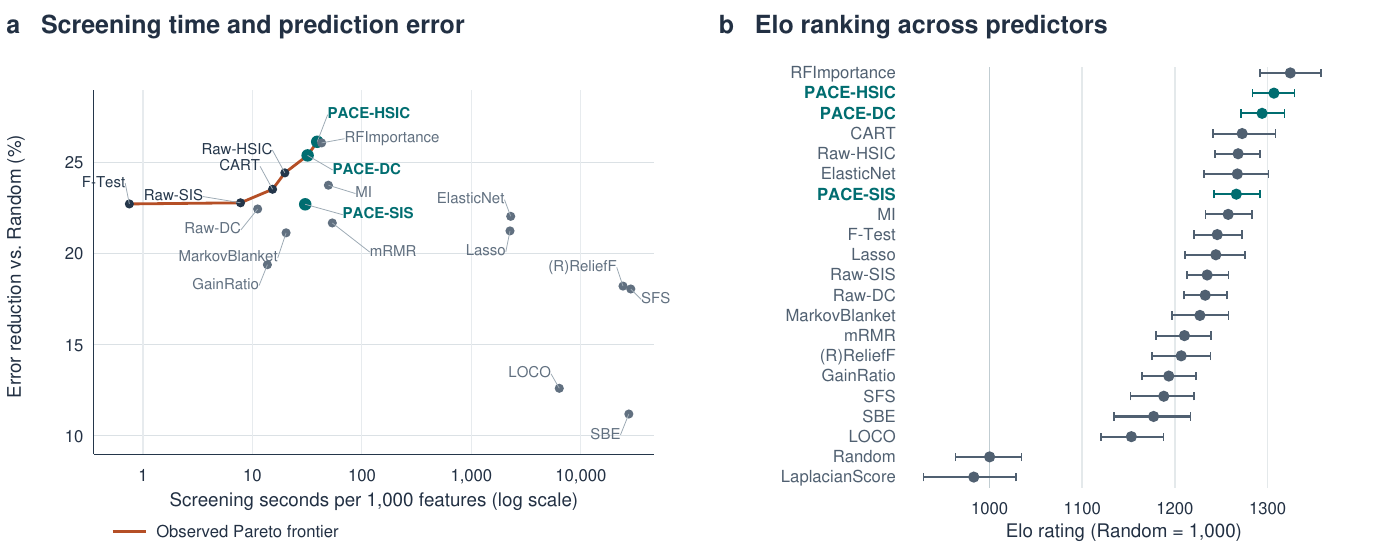}
\caption{\textbf{SelectArena performance.} (a) Error reduction from Random versus screening time; the line marks the observed frontier.  (b) Elo ratings with 95\% bootstrap intervals (Random: 1,000).}
\label{fig:selectarena_comparison}
\end{figure}

{As shown in Figure~\ref{fig:selectarena_comparison}, all three PACE variants achieve higher Elo ratings than their raw counterparts. PACE-HSIC and PACE-DC rank second and third overall, respectively, and both lie on the observed performance--time Pareto frontier. Table~\ref{tab:selectarena_class_budgets} further examines their performance across the five feature budgets on the 61 classification datasets. PACE-HSIC achieves the best performance at the two smallest budgets, while PACE-DC consistently improves over raw DC across all five budgets. In terms of the five-budget mean error reduction, PACE-DC improves from 21.00\% to 24.15\%, and PACE-HSIC from 22.87\% to 24.62\%, corresponding to gains of 3.15 and 1.75 percentage points, respectively. PACE-SIS also achieves a higher overall mean than raw SIS (21.72\% versus 21.18\%), although its improvement is more budget-dependent. These results demonstrate that pretrained contextual representations can improve the predictive utility of classical screening methods on natural datasets, with particularly consistent benefits for DC- and HSIC-based screening.}

\begin{table}[t]
\centering
\caption{\textbf{Classification error reduction from Random (\%).} Budgets increase from left to right; Mean averages the five columns equally. Bold: best among all 21 selectors.}
\label{tab:selectarena_class_budgets}
\small
\begin{tabular}{lrrrrrr}
\toprule
Selector & Budget 1~(low) & Budget 2 & Budget 3 & Budget 4 & Budget 5~(high) & Mean \\
\midrule
Raw-SIS & 31.12 & 28.68 & 20.37 & 14.90 & 10.82 & 21.18 \\
PACE-SIS & 28.14 & 29.68 & 22.44 & 16.72 & 11.60 & 21.72 \\
Raw-DC & 31.29 & 27.77 & 21.37 & 14.89 & 9.71 & 21.00 \\
PACE-DC & 33.36 & 32.36 & 25.18 & 17.77 & 12.05 & 24.15 \\
Raw-HSIC & 33.61 & 31.25 & 23.36 & 15.56 & 10.57 & 22.87 \\
PACE-HSIC & \textbf{35.40} & \textbf{33.36} & 25.35 & 17.22 & 11.79 & \textbf{24.62} \\
RFImportance & 32.55 & 31.51 & \textbf{26.45} & \textbf{19.73} & \textbf{12.71} & 24.59 \\
\bottomrule
\end{tabular}
\end{table}

\subsection{Representation ablations: backbone, pretraining, and extraction}
\label{sec:ablation}

\begin{figure}[H]
\centering
\includegraphics[width=.9\linewidth]{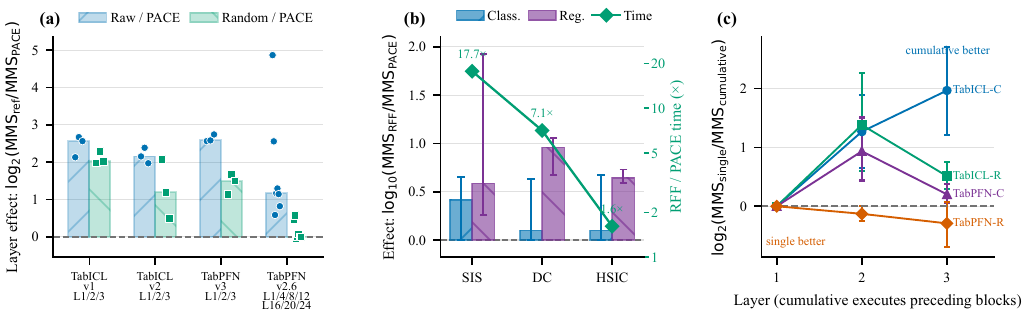}
\caption{\textbf{Representation ablations.} (a) Backbone and random-weight controls: points denote layers and bars their medians. (b) RFF comparison: positive values favor PACE and labels give RFF-to-PACE time ratios. (c) Cumulative versus same-layer single-block extraction: C/R denote classification/regression.}
\label{fig:ablation_summary}
\end{figure}

Across controlled settings, Figure~\ref{fig:ablation_summary} examines whether the screening gains arise from the pretrained representation itself and how they depend on the extraction choice. We summarize each variant by first aggregating paired log-MMS ratios across utilities within a setting and then taking the median across settings; positive values indicate that fewer top-ranked features are required to recover the active set.

In panel~(a), for the main TabICL v1-$L_2$ representation, the resulting effects correspond to MMS reduction factors of $6.38$ relative to raw input and $4.91$ relative to its matched random-weight counterpart. The latter control preserves the Transformer architecture, representation dimension, and extraction procedure while replacing pretrained weights with random initialization, thereby isolating the contribution of pretraining from architectural expansion. Positive contrasts for TabICL v2 and TabPFN v3 further show that the gain extends beyond the main TabICL v1 backbone.

Panel~(b) provides a complementary nonlinear-expansion control using RFF with the same output dimension as the main TabICL representation. PACE has lower aggregate MMS than RFF in all six task--utility summaries, while the panel also reports their relative screening costs. The RFF map uses random features with an input-dependent bandwidth. Together with the matched random-weight comparison, these results provide evidence that dimensional expansion alone does not explain the observed gains; the structure learned during TFM pretraining contribute additional screening value.

Finally, panel~(c) separates cumulative extraction, which executes all preceding blocks, from applying the same target block in isolation. At $L_2/L_3$, cumulative execution lowers MMS for TabICL v2 and for TabPFN v3 classification, whereas single-block extraction tends to perform better for TabPFN v3 regression. The preferred extraction path varies across backbones and tasks, showing that both the pretrained representation and its extraction affect screening performance. Appendix Figures~\ref{fig:mechanism_main}--\ref{fig:ablation_utility} provide the full backbone-, layer-, and utility-level results.

\section{Conclusion}
\label{sec:limitations}
PACE provides a reusable feature-screening representation adapter using a frozen TFM column encoder. By preserving original feature identity while changing its representation, PACE helps existing utilities expose complex nonlinear and distributional signals while remaining compatible with diverse downstream predictors. Across controlled and real-data evaluations, the gains extend across utilities and learner families. PACE provides favorable performance--time trade-offs on wide tables, and ablations show that pretrained structure contributes beyond generic dimensional expansion.

Screening performance still depends on the representation, utility, backbone, and extraction strategy. Future work includes combining raw and pretrained representations, exploring additional pretrained components, and reducing the quadratic cost in sample size of distance- and kernel-based utilities. Because column-wise attention can induce dependence across embedded observations, establishing sure screening properties under this dependence is another important direction. More broadly, our results suggest that pretrained tabular representations can serve not only predictive models but also reusable upstream statistical procedures.

\clearpage
\bibliography{references}
\bibliographystyle{preprint}

\clearpage
\appendix
\input{appendix_main}

\end{document}

%% file: appendix_main.tex
\section{Appendix Overview}
\label{app:overview}

The appendix is organized as follows. Appendix~\ref{app:algorithm} gives the screening procedure, with utility formulas in Appendix~\ref{app:utility_formulas}. Appendix~\ref{app:controlled} presents complete matched raw-to-PACE simulation results, implementation details for the cross-paradigm baselines, and setting-level MMS comparisons. Appendix~\ref{app:formal} proves the information-boundary result and states its scope. Appendix~\ref{app:real_full} presents the full nuisance-contaminated TALENT evaluation, including matched comparisons across downstream learners, model-family sensitivity, contamination recovery, a broad screening comparison, and natural high-dimensional classification. Appendix~\ref{app:ablation_full} extends the main-text ablation analysis across pretrained backbones, extraction layers, screening utilities, and mean versus maximum coordinate aggregation. Appendix~\ref{app:repro} provides implementation, model, data, and responsible-use details.

Collectively, the empirical study covers 25 controlled settings; 288 nuisance-contaminated TALENT datasets; ten natural high-dimensional classification datasets evaluated with TabPFN v3~\citep{grinsztajn2026tabpfn} and TabICL v2~\citep{qu2026tabiclv2} using classical, learned, attribution-based, and tree-importance selectors, as well as with TabPFN-Wide~\citep{kolberg2025tabpfn} and GOTabPFN~\citep{habib2026gotabpfn}; and representation analyses across multiple backbones, layers, and screening utilities.  Additional controls include five supervised learned selectors and matched pretrained and random-weight representations.

\section{Algorithmic Specification}
\label{app:algorithm}

\begin{algorithm}[H]
\caption{\method{} screening with a response-blind column encoder.}
\label{alg:pace}
\begin{algorithmic}[1]
\Require Training feature columns $\{X_j\}_{j=1}^p$ and response $Y$; frozen column encoder $G_{\theta,N}$
\Require Prespecified screening utility $\mathcal U$; selection budget $k$
\Ensure Scores $\widehat{\boldsymbol\omega}\in\R^p$ and selected original-feature indices $\widehat{\mathcal S}_k$
\For{$j=1,\ldots,p$ \textbf{in parallel}}
    \State $\widetilde X_j\gets\textsc{Preprocess}(X_j)$ under the fixed study protocol
    \State $\mathbf H_j\gets G_{\theta,N}(\widetilde X_j)$ \Comment{$Y$ is not passed to $G_{\theta,N}$}
    \If{$\mathcal U$ accepts vector-valued input}
        \State $\widehat\omega_j\gets\mathcal U(\mathbf H_j,Y)$
    \Else
        \For{$m=1,\ldots,d$}
            \State $u_{jm}\gets\mathcal U(\mathbf H_j^{(:,m)},Y)$
        \EndFor
        \State $\widehat\omega_j\gets d^{-1}\sum_{m=1}^d u_{jm}$
    \EndIf
\EndFor
\State $\pi\gets\operatorname{argsort}_{j=1,\ldots,p}(\widehat\omega_j)$ in decreasing order
\State $\widehat{\mathcal S}_k\gets\{\pi_1,\ldots,\pi_k\}$
\State \Return $\widehat{\boldsymbol\omega}$ and $\widehat{\mathcal S}_k$
\end{algorithmic}
\end{algorithm}

The encoder $G_{\theta,N}$ uses feature values only, while the supervised utility $\mathcal U(\cdot,Y)$ also receives the training response. Appendix~\ref{app:real_protocol} describes the real-data preprocessing.

\begin{figure}[!htbp]
\centering
\includegraphics[width=\linewidth]{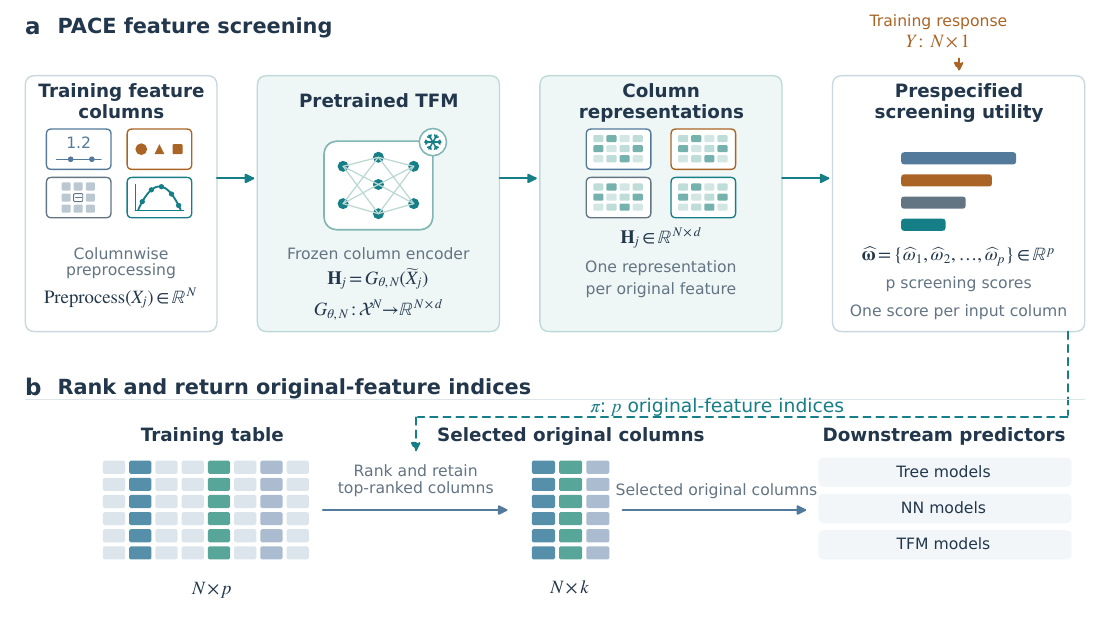}
\caption{{\textbf{PACE workflow (Algorithm~\ref{alg:pace}).} (a) A frozen TFM column encoder produces response-blind representations; the screening utility assigns one score per original feature. (b) The top-$k$ original columns are retained for downstream prediction.}}
\label{fig:pace_workflow}
\end{figure}

\subsection{Screening utility definitions}
\label{app:utility_formulas}

We define the screening utilities using a common notation. Let $Z=(z_1,\ldots,z_N)^\top\in\R^{N\times q}$ denote the input for one original feature: $q=1$ for raw screening and $q=d$ for its PACE representation. For dependence measures with vector inputs, use the response representation $V=\psi(Y)$ defined in the main text, with rows $v_i^\top$.

For a scalar-valued feature $z$ and a classification response, let $\mathcal I_r=\{i:y_i=r\}$, $N_r=|\mathcal I_r|$, and $\widehat\pi_r=N_r/N$, with $r=1,\ldots,R$ indexing the observed classes. Define the empirical marginal, class-conditional, and complementary-class distribution functions by
\begin{equation}
\label{eq:utility_ecdf}
\begin{aligned}
\widehat F(t)&=\frac{1}{N}\sum_{i=1}^N\mathbb I(z_i\le t),&
\widehat F_r(t)&=\frac{1}{N_r}\sum_{i\in\mathcal I_r}\mathbb I(z_i\le t),\\
\widehat F_{-r}(t)&=\frac{1}{N-N_r}\sum_{i\notin\mathcal I_r}\mathbb I(z_i\le t).
\end{aligned}
\end{equation}
We use $C_N=I_N-N^{-1}\boldsymbol 1\boldsymbol 1^\top$ for centering and $\langle A,B\rangle_F=\sum_{i,\ell}A_{i\ell}B_{i\ell}$ for the Frobenius inner product, with norm $\|A\|_F=\sqrt{\langle A,A\rangle_F}$. Normalized scores are set to zero when their denominator vanishes.

\paragraph{SIS.}
SIS~\citep{fan2008sure} uses absolute marginal Pearson correlation. For vectors $a,b\in\R^N$ with sample means $\bar a$ and $\bar b$, define
\begin{equation}
\label{eq:utility_corr}
\widehat\rho(a,b)=
\frac{\sum_{i=1}^N(a_i-\bar a)(b_i-\bar b)}
{\left\{\sum_{i=1}^N(a_i-\bar a)^2\sum_{i=1}^N(b_i-\bar b)^2\right\}^{1/2}}.
\end{equation}
The regression score is $\mathcal U_{\rm SIS}(z,Y)=|\widehat\rho(z,Y)|$. For classification, we use the largest absolute one-versus-rest correlation,
\begin{equation}
\label{eq:utility_sis}
\mathcal U_{\rm SIS}(z,Y)=\max_{1\le r\le R}
\left|\widehat\rho\bigl(z,\mathbb I(Y=r)\bigr)\right|.
\end{equation}

\paragraph{DC.}
Distance-correlation screening~\citep{li2012feature} operates on pairwise Euclidean distances. Let $D^Z_{i\ell}=\|z_i-z_\ell\|_2$, $D^V_{i\ell}=\|v_i-v_\ell\|_2$, $A=C_ND^ZC_N$, and $B=C_ND^VC_N$. The sample distance correlation is
\begin{equation}
\label{eq:utility_dc}
\mathcal U_{\rm DC}(Z,Y)=
\left\{\frac{\langle A,B\rangle_F}
{\|A\|_F\|B\|_F}\right\}^{1/2}.
\end{equation}
Thus the distances are computed between scalar observations for raw screening and between full embedding vectors for PACE.

\paragraph{FKF.}
The Kolmogorov-filter family~\citep{mai2015fused} compares conditional distribution functions. The classification implementation uses the largest one-versus-rest Kolmogorov discrepancy,
\begin{equation}
\label{eq:utility_fkf}
\mathcal U_{\rm FKF}(z,Y)=
\max_{1\le r\le R}\ \sup_{t\in\R}
\left|\widehat F_r(t)-\widehat F_{-r}(t)\right|.
\end{equation}
The supremum is evaluated over the pooled observed feature values.

\paragraph{MV.}
The mean--variance index of \citet{cui2015model} measures variation in the conditional distribution function across response classes:
\begin{equation}
\label{eq:utility_mv_population}
\operatorname{MV}(z\mid Y)
=\sum_{r=1}^R\pi_r\int\{F_r(t)-F(t)\}^2\,dF(t).
\end{equation}
Here $F$, $F_r$, and $\pi_r$ are the population counterparts of the quantities in Equation~\eqref{eq:utility_ecdf}. The empirical counterpart of the mean--variance index is
\begin{equation}
\label{eq:utility_mv}
\mathcal U_{\rm MV}(z,Y)
=\frac{1}{N}\sum_{i=1}^N\sum_{r=1}^R
\widehat\pi_r\left\{\widehat F_r(z_i)-\widehat F(z_i)\right\}^2.
\end{equation}

\paragraph{PSIS.}
Pairwise screening~\citep{pan2016ultrahigh} compares feature locations across response classes. Write $\widehat\mu_r=N_r^{-1}\sum_{i\in\mathcal I_r}z_i$ and $\widehat m_r=\operatorname{median}\{z_i:i\in\mathcal I_r\}$. The corresponding mean- and median-contrast scores are
\begin{equation}
\label{eq:utility_psis}
\mathcal U_{\rm PSIS}^{\rm mean}(z,Y)=
\max_{r<s}|\widehat\mu_r-\widehat\mu_s|,
\qquad
\mathcal U_{\rm PSIS}^{\rm median}(z,Y)=
\max_{r<s}|\widehat m_r-\widehat m_s|.
\end{equation}
The original method uses mean contrasts; our implementation uses the median-contrast variant, with the lower middle order statistic when a class has an even sample size.

\paragraph{CAVS.}
Category-adaptive screening~\citep{xie2020category} defines a class-specific utility for a continuous scalar feature $X$ as
\begin{equation}
\label{eq:utility_cavs_population}
\tau_r=\mathbb E_X\{F_r(X)\}-\frac12
=\int F_r(t)\,dF(t)-\frac12.
\end{equation}
The expectation is taken over the marginal feature distribution. The original sample estimator is
\begin{equation}
\label{eq:utility_cavs_class}
\widehat\tau_r=
\frac{1}{N+1}\sum_{k=1}^N\widehat F_r(z_k)-\frac12
=\frac{1}{(N+1)N_r}
\sum_{k=1}^N\sum_{i\in\mathcal I_r}\mathbb I(z_i\le z_k)-\frac12.
\end{equation}
Aggregating the class-specific contrasts gives
\begin{equation}
\label{eq:utility_cavs}
\mathcal U_{\rm CAVS}(z,Y)=
\max_{1\le r\le R}|\widehat\tau_r|.
\end{equation}
The $N+1$ denominator makes each class-specific estimator mean-zero under independence for continuous features.

\paragraph{HSIC.}
HSIC-based screening~\citep{poignard2022feature} uses the biased empirical Hilbert--Schmidt independence criterion
\begin{equation}
\label{eq:utility_hsic}
\mathcal U_{\rm HSIC}(Z,Y)=
\frac{1}{N^2}\operatorname{tr}(K C_N L C_N),
\qquad
K_{i\ell}=\exp\left(-\frac{\|z_i-z_\ell\|_2^2}{2\sigma_Z^2}\right).
\end{equation}
For classification, $L_{i\ell}=\mathbb I(y_i=y_\ell)$; for regression, $L_{i\ell}=\exp\{-(y_i-y_\ell)^2/(2\sigma_Y^2)\}$. The bandwidth for each RBF kernel is the median of its nonzero pairwise distances, computed from the training observations. A constant input receives a score of zero. PACE uses the full embedding vector when constructing $K$.

\paragraph{PC.}
The original sample projection correlation~\citep{liu2022model} averages angles around every sample anchor. For each anchor $t$, define the $N\times N$ angle matrix
\begin{equation}
\label{eq:utility_pc_angles}
a^{Z,t}_{i\ell}=
\arccos\left(
\frac{(z_i-z_t)^\top(z_\ell-z_t)}
{\|z_i-z_t\|_2\|z_\ell-z_t\|_2}
\right),
\qquad
A^{Z,t}=C_Na^{Z,t}C_N,
\end{equation}
with the angle set to zero whenever either difference vector vanishes. Define $A^{V,t}$ in the same way from $V$, and let
\begin{equation}
\label{eq:utility_pcov}
S_{\rm P}(Z,V)=\frac{1}{N^3}
\sum_{t=1}^N\langle A^{Z,t},A^{V,t}\rangle_F.
\end{equation}
The projection-correlation utility is
\begin{equation}
\label{eq:utility_pc}
\mathcal U_{\rm PC}(Z,Y)=
\left\{\frac{S_{\rm P}(Z,V)}
{\sqrt{S_{\rm P}(Z,Z)S_{\rm P}(V,V)}}\right\}^{1/2}.
\end{equation}
The squared version induces the same feature ranking.

\paragraph{HD.}
Hellinger-distance screening~\citep{wu2024model} compares both the conditional CDF and its complement:
\begin{equation}
\label{eq:utility_hd}
\begin{aligned}
\mathcal U_{\rm HD}(z,Y)=\frac{1}{N}\sum_{i=1}^N\sum_{r=1}^R\widehat\pi_r
\Bigl[&
\left\{\sqrt{\widehat F_r(z_i)}-\sqrt{\widehat F(z_i)}\right\}^2\\
&+\left\{\sqrt{1-\widehat F_r(z_i)}-\sqrt{1-\widehat F(z_i)}\right\}^2
\Bigr].
\end{aligned}
\end{equation}
For regression, let $Y^{(\ell)}$ be the categorical response obtained by partitioning $Y$ at empirical quantiles into $\ell+3$ slices. The implementation averages the resulting classification scores:
\begin{equation}
\label{eq:utility_hd_reg}
\mathcal U_{\rm HD}^{\rm reg}(z,Y)=
\frac{1}{L}\sum_{\ell=1}^L\mathcal U_{\rm HD}(z,Y^{(\ell)}),
\qquad
L=\max\{1,\lceil\log N\rceil-2\}.
\end{equation}
Empty slices contribute zero. The CDFs are computed cumulatively over the sorted feature observations.

\paragraph{CRU.}
Conditional-rank screening~\citep{li2024feature} measures class-specific deviations in mean marginal rank. For a continuous scalar feature $X$ with CDF $F$, its population utility can be written as
\begin{equation}
\label{eq:utility_cru_population}
\begin{aligned}
\operatorname{CRU}(X,Y)
&=\sum_{r=1}^R
\left[\mathbb E\{F(X)\mathbb I(Y=r)\}-\frac{\pi_r}{2}\right]^2\\
&=\sum_{r=1}^R\pi_r^2
\left[\mathbb E\{F(X)\mid Y=r\}-\frac12\right]^2.
\end{aligned}
\end{equation}
For pooled observations, the component $\theta_r=\mathbb E\{F(X)\mathbb I(Y=r)\}$ has the unbiased pairwise estimator
\begin{equation}
\label{eq:utility_cru_components}
\widehat\theta_r=
\frac{1}{N(N-1)}\sum_{i=1}^N\sum_{\ell\ne i}
\mathbb I(z_\ell\le z_i)\mathbb I(y_i=r).
\end{equation}
Substitution gives the empirical score
\begin{equation}
\label{eq:utility_cru}
\mathcal U_{\rm CRU}(z,Y)=
\sum_{r=1}^R\left(\widehat\theta_r-\frac{\widehat\pi_r}{2}\right)^2.
\end{equation}

\paragraph{GCor.}
The GCor score~\citep{jiang2025robust} is constructed from the Grothendieck kernel
\begin{equation}
\label{eq:utility_gcor_kernel}
K^Z_{i\ell}=
\arcsin\left(
\frac{1+z_i^\top z_\ell}
{\sqrt{(1+\|z_i\|_2^2)(1+\|z_\ell\|_2^2)}}
\right).
\end{equation}
For $N>3$, first set the diagonal of $K^Z$ to zero, obtaining $K^{Z,0}$, and apply $U$-centering:
\begin{equation}
\label{eq:utility_ucenter}
\begin{aligned}
\widetilde K^Z_{i\ell}
={}&K^{Z,0}_{i\ell}
-\frac{\sum_bK^{Z,0}_{ib}}{N-2}
-\frac{\sum_aK^{Z,0}_{a\ell}}{N-2}
+\frac{\sum_{a,b}K^{Z,0}_{ab}}{(N-1)(N-2)},
\quad i\ne\ell,\\
\widetilde K^Z_{ii}={}&0.
\end{aligned}
\end{equation}
Construct $\widetilde K^V$ analogously and set
\begin{equation}
\label{eq:utility_gcor}
S_{\rm G}(Z,V)=
\frac{\langle\widetilde K^Z,\widetilde K^V\rangle_F}{N(N-3)},
\qquad
\mathcal U_{\rm GCor}(Z,Y)=
\left\{\frac{|S_{\rm G}(Z,V)|}
{\sqrt{S_{\rm G}(Z,Z)S_{\rm G}(V,V)}}\right\}^{1/2}.
\end{equation}
The absolute value specifies the implemented treatment of a negative finite-sample cross-covariance estimate.

\paragraph{LR-FFS.}
The pooled-data LR-FFS score~\citep{qin2025label} uses the largest unweighted class-specific rank contrast. Define the empirical pairwise rank probability using strict comparisons:
\begin{equation}
\label{eq:utility_gamma}
\widehat\gamma_r(z)=
\frac{1}{N_r(N-N_r)}
\sum_{i\in\mathcal I_r}\sum_{\ell\notin\mathcal I_r}
\mathbb I(z_i<z_\ell).
\end{equation}
The score is
\begin{equation}
\label{eq:utility_lrffs}
\mathcal U_{\rm LR\text{-}FFS}(z,Y)=
\max_{1\le r\le R}\left|\widehat\gamma_r(z)-\frac12\right|,
\end{equation}
with $\widehat\gamma_r$ as in Equation~\eqref{eq:utility_gamma}. 

\paragraph{IPC.}
For IPC~\citep{lin2026model}, the implemented transformed-Gram score first standardizes each input coordinate. Let $\widetilde z_i$ denote a row after centering and division by its coordinatewise sample standard deviation. Define
\begin{equation}
\label{eq:utility_ipc_map}
\sigma_Z^2=\operatorname{median}_{i}\|\widetilde z_i\|_2^2,
\qquad
g_i^Z=\frac{\widetilde z_i}
{\sigma_Z\sqrt{\sigma_Z^2+\|\widetilde z_i\|_2^2}},
\qquad
M_Z=G_ZG_Z^\top,
\end{equation}
where $G_Z$ has rows $(g_i^Z)^\top$. Constant coordinates are standardized to zero. Apply the same construction to $V$ to obtain $M_V$, assigning a score of zero if either scale parameter vanishes. Otherwise, the score is the normalized centered Gram-matrix inner product
\begin{equation}
\label{eq:utility_ipc}
\mathcal U_{\rm IPC}(Z,Y)=
\frac{\langle C_NM_ZC_N,C_NM_VC_N\rangle_F}
{\|C_NM_ZC_N\|_F\|C_NM_VC_N\|_F}.
\end{equation}

\paragraph{Raw and PACE inputs.}
DC, HSIC, PC, GCor, and IPC admit the vector-valued input $Z=\mathbf H_j$ directly. For the scalar utilities SIS, FKF, MV, PSIS, CAVS, HD, CRU, and LR-FFS, the PACE interface applies the utility to each embedding coordinate and averages the resulting scores:
\begin{equation}
\label{eq:utility_pace_aggregation}
\widehat\omega_j^{\rm raw}=\mathcal U(X_j,Y),
\qquad
\widehat\omega_j^{\rm PACE}=
\begin{cases}
\mathcal U(\mathbf H_j,Y), & \text{vector-input utility},\\
d^{-1}\sum_{m=1}^d\mathcal U(\mathbf H_j^{(:,m)},Y), & \text{scalar-input utility}.
\end{cases}
\end{equation}
Larger scores receive higher ranks, and the selected indices always refer to original feature columns.

\section{Controlled Simulation Studies}
\label{app:controlled}

\subsection{Raw and PACE versions of feature-screening utilities}
\label{app:raw_pace_simulation}

The controlled raw/PACE evaluation contains 25 settings. Within each raw/PACE comparison, a replication evaluates the paired utilities on the same generated dataset. The known target support enters the post-ranking MMS calculation. Regular raw/PACE settings use 50 replications. Ultrahigh-dimensional settings use 25 replications.

The primary analysis presents classification utilities in the chronological order of their cited methodological sources: SIS~\citep{fan2008sure}, DC~\citep{li2012feature}, FKF~\citep{mai2015fused}, MV~\citep{cui2015model}, PSIS~\citep{pan2016ultrahigh}, CAVS~\citep{xie2020category}, HSIC~\citep{poignard2022feature}, PC~\citep{liu2022model}, HD~\citep{wu2024model}, CRU~\citep{li2024feature}, GCor~\citep{jiang2025robust}, LR-FFS~\citep{qin2025label}, and IPC~\citep{lin2026model}. The regression set contains SIS, DC, HSIC, PC, HD, GCor, and IPC in the same order. The matched design compares each utility on raw values and pretrained representations under the same screening budget.

\subsubsection{Classification data-generating processes}

The 16 classification settings instantiate six design families.

\paragraph{exp1: strong additive classification ($N=200$, $p=3000$, $s=8$).}
This design tests whether screening preserves readily detectable class mean-shift signals across different feature distributions.
Draw $Y\sim\operatorname{Bernoulli}(1/2)$ and let $\mathcal S=\{1,\ldots,8\}$.  Active features receive opposing class shifts of magnitude $0.6$; inactive features have identical class-conditional distributions.  The three variants use Gaussian, $t_3$, and log-normal base distributions, respectively.

\paragraph{exp2: nonlinear latent classification ($N=300$, $p=3000$, $s=4$).}
This design tests recovery of nonlinear signals combining linear and quadratic effects under different feature correlations and noise distributions.
Draw $\boldsymbol X\sim\mathcal N(0,\Sigma)$ and
\[
Y^*=2X_1+2X_2+2.5(X_3^2-1)+2.5(X_4^2-1)+3\varepsilon,
\qquad Y=\mathbb I(Y^*>0).
\]
The three variants use identity covariance with Gaussian errors; AR(1) covariance, $\Sigma_{ab}=0.3^{|a-b|}$, with $t_3$ errors; and banded covariance with adjacent correlation $0.3$ and Gaussian errors, respectively.

\paragraph{exp3: three-class bimodal signal ($N=300$, $p=2000$, $s=10$).}
This design tests recovery of class-dependent distributional signals with equal class means under varying class imbalance.
Features 1--5 carry a class-1 signal and features 6--10 carry a class-2 signal.  For each active feature, observations from its signal-bearing class follow $0.5\mathcal N(-2,0.25)+0.5\mathcal N(2,0.25)$, whereas observations from the remaining classes and all inactive features follow $\mathcal N(0,1)$.  The three variants use class probabilities $(0.70,0.20,0.10)$, $(0.85,0.10,0.05)$, and $(0.33,0.34,0.33)$, respectively.

\paragraph{exp4: ultrahigh-dimensional classification ($N=400$, $s=15$).}
This design tests recovery of mean- and scale-shift signals as the number of nuisance features increases.
For 12 active features, the class-conditional distributions are $\mathcal N(-4.5,1)$ and $\mathcal N(4.5,1)$.  Three additional active features have equal means but class-dependent scales, $\mathcal N(0,1)$ versus $\mathcal N(0,2.5^2)$.  All inactive features are standard Gaussian, and $p\in\{10\mathrm K,50\mathrm K,100\mathrm K\}$.

\paragraph{exp5: mixed continuous--categorical classification ($N=300$, $p=2000$, $s=6$).}
This design tests screening of mixed continuous and categorical signals as the number of categorical levels varies.
Three active continuous features have opposite class means.  Three active categorical features take $K$ levels, are uniform in class 0, and follow a nonmonotone reweighting over a random subset of levels in class 1.  The two variants use $(K,\mu,c)=(25,0.8,4)$ and $(40,0.6,5)$, respectively; inactive features mix Gaussian and categorical noise.

\paragraph{exp6: missing feature values ($N=300$, $p=3000$, $s=4$).}
This design tests nonlinear signal recovery with correlated features and increasing rates of missing values.
These two variants use the latent response form of exp2 with AR(1) covariance, $\Sigma_{ab}=0.3^{|a-b|}$, and independently mask each feature value with probabilities $0.15$ and $0.30$, respectively.
\subsubsection{Full classification results}
\label{app:class_full}

\begin{table}[!htbp]
\centering
\caption{\textbf{Controlled classification: raw versus PACE screening.} Thirteen utilities span exp1--exp6, with parameter sets ordered within each experiment. Panels (a) and (b) apply each utility to raw features and PACE-$L_2$ representations, respectively. Cells report median MMS (IQR); lower is better.}
\label{tab:class_full}
\scriptsize
\setlength{\tabcolsep}{2.5pt}
\resizebox{\textwidth}{!}{%
\begin{tabular}{lrrrrrrrrrrrrrr}
\toprule
\multicolumn{15}{l}{(a) Raw feature screening}\\
Setting & $s$ & SIS & DC & FKF & MV & PSIS & CAVS & HSIC & PC & HD & CRU & GCor & LR-FFS & IPC\\
\midrule
exp1/1 (Gaussian) & 8 & 8 (0) & 8 (0) & 8 (0) & 8 (0) & 8 (0) & 8 (0) & 8 (0) & 8 (0) & 8 (0) & 8 (0) & 8 (0) & 8 (0) & 8 (0)\\
exp1/2 ($t_3$) & 8 & 9 (22) & 8 (0) & 8 (0) & 8 (0) & 8 (0) & 8 (0) & 8 (0) & 8 (0) & 8 (0) & 8 (0) & 8 (0) & 8 (0) & 8 (0)\\
exp1/3 (log-normal) & 8 & 8 (0) & 8 (0) & 8 (0) & 8 (0) & 8 (0) & 8 (0) & 8 (0) & 8 (0) & 8 (0) & 8 (0) & 8 (0) & 8 (0) & 8 (0)\\
exp2/1 (independent) & 4 & 2179 (1063) & 6 (4) & 21 (34) & 2208 (1146) & 1782 (1374) & 2208 (1146) & 4 (2) & 18 (22) & 4 (1) & 2204 (1154) & 18 (22) & 2209 (1146) & 2053 (1116)\\
exp2/2 (AR(1)/$t_3$) & 4 & 1736 (1306) & 6 (4) & 28 (40) & 1827 (1446) & 1907 (1331) & 1825 (1444) & 4 (1) & 23 (26) & 4 (2) & 1824 (1441) & 21 (27) & 1824 (1446) & 1872 (1109)\\
exp2/3 (banded) & 4 & 1664 (1368) & 5 (1) & 18 (24) & 1668 (1331) & 1628 (1335) & 1668 (1331) & 4 (0) & 10 (10) & 4 (0) & 1668 (1336) & 12 (10) & 1668 (1330) & 1671 (1347)\\
exp3/1 (moderate imbalance) & 10 & 1784 (213) & 16 (19) & 12 (4) & 1842 (272) & 13 (18) & 1836 (236) & 10 (1) & 72 (109) & 12 (4) & 1843 (272) & 75 (118) & 1826 (224) & 1828 (256)\\
exp3/2 (severe imbalance) & 10 & 1728 (256) & 86 (114) & 45 (74) & 1792 (263) & 48 (96) & 1774 (200) & 26 (31) & 278 (294) & 86 (128) & 1791 (265) & 833 (1378) & 1782 (219) & 1785 (312)\\
exp3/3 (balanced) & 10 & 1868 (167) & 10 (0) & 10 (0) & 1840 (174) & 10 (1) & 1845 (160) & 10 (0) & 10 (0) & 10 (0) & 1836 (175) & 10 (0) & 1840 (177) & 1836 (216)\\
exp4/1 (10K features) & 15 & 7826 (2818) & 15 (0) & 15 (1) & 7002 (2489) & 4653 (4359) & 7002 (2489) & 15 (0) & 15 (0) & 15 (0) & 7006 (2496) & 15 (2) & 7000 (2488) & 5388 (3716)\\
exp4/2 (50K features) & 15 & 32136 (9411) & 15 (0) & 16 (3) & 34780 (13380) & 26131 (21731) & 34774 (13392) & 15 (0) & 15 (1) & 15 (0) & 34840 (13326) & 22 (12) & 34801 (13394) & 30811 (16266)\\
exp4/3 (100K features) & 15 & 86096 (28678) & 15 (0) & 22 (16) & 87210 (12109) & 84574 (27024) & 87207 (12109) & 15 (0) & 15 (1) & 15 (0) & 87186 (11906) & 24 (14) & 87199 (12038) & 79168 (14886)\\
exp5/2 ($K=25$) & 6 & 753 (1455) & 30 (96) & 12 (48) & 654 (1186) & 471 (1460) & 568 (1144) & 118 (569) & 75 (304) & 28 (130) & 752 (1425) & 89 (135) & 516 (1074) & 696 (1458)\\
exp5/3 ($K=40$) & 6 & 898 (1144) & 85 (234) & 60 (202) & 884 (1165) & 726 (290) & 744 (1164) & 292 (992) & 272 (663) & 82 (298) & 894 (1330) & 144 (1208) & 746 (1159) & 982 (1251)\\
exp6/1 (MCAR 15\%) & 4 & 1991 (1296) & 7 (7) & 1200 (971) & 1752 (1054) & 3000 (0) & 1643 (1302) & 4 (1) & 31 (62) & 6 (6) & 1908 (1136) & 30 (55) & 1810 (1116) & 1708 (1245)\\
exp6/2 (MCAR 30\%) & 4 & 2128 (1046) & 20 (48) & 2010 (1125) & 1860 (1300) & 3000 (0) & 1818 (1129) & 7 (25) & 130 (224) & 28 (51) & 2112 (1063) & 86 (181) & 2018 (1034) & 2061 (1454)\\
\bottomrule
\end{tabular}}
\vspace{.5em}
\resizebox{\textwidth}{!}{%
\begin{tabular}{lrrrrrrrrrrrrrr}
\toprule
\multicolumn{15}{l}{(b) Matched PACE-$L_2$ feature screening}\\
Setting & $s$ & SIS & DC & FKF & MV & PSIS & CAVS & HSIC & PC & HD & CRU & GCor & LR-FFS & IPC\\
\midrule
exp1/1 (Gaussian) & 8 & 8 (0) & 8 (0) & 8 (0) & 8 (0) & 8 (0) & 8 (0) & 8 (0) & 8 (0) & 8 (0) & 8 (0) & 8 (0) & 8 (0) & 8 (0)\\
exp1/2 ($t_3$) & 8 & 8 (0) & 8 (0) & 8 (0) & 8 (0) & 8 (0) & 8 (0) & 8 (0) & 8 (0) & 8 (0) & 8 (0) & 8 (0) & 8 (0) & 8 (0)\\
exp1/3 (log-normal) & 8 & 8 (0) & 8 (0) & 8 (0) & 8 (0) & 8 (0) & 8 (0) & 8 (0) & 8 (0) & 8 (0) & 8 (0) & 8 (0) & 8 (0) & 8 (0)\\
exp2/1 (independent) & 4 & 4 (2) & 4 (1) & 8 (10) & 6 (5) & 148 (143) & 9 (16) & 5 (4) & 7 (11) & 4 (2) & 6 (5) & 4 (2) & 9 (16) & 6 (6)\\
exp2/2 (AR(1)/$t_3$) & 4 & 4 (1) & 4 (0) & 5 (5) & 5 (4) & 130 (175) & 11 (14) & 4 (1) & 6 (5) & 4 (1) & 5 (4) & 4 (1) & 11 (14) & 5 (3)\\
exp2/3 (banded) & 4 & 4 (0) & 4 (0) & 4 (1) & 4 (1) & 66 (90) & 6 (5) & 4 (0) & 4 (2) & 4 (0) & 4 (1) & 4 (0) & 6 (5) & 4 (1)\\
exp3/1 (moderate imbalance) & 10 & 10 (0) & 10 (1) & 10 (0) & 10 (0) & 10 (0) & 10 (2) & 10 (0) & 12 (11) & 10 (0) & 10 (0) & 10 (0) & 10 (0) & 16 (16)\\
exp3/2 (severe imbalance) & 10 & 10 (1) & 21 (30) & 10 (2) & 12 (6) & 11 (5) & 16 (11) & 18 (15) & 72 (89) & 10 (2) & 12 (6) & 16 (28) & 16 (11) & 84 (126)\\
exp3/3 (balanced) & 10 & 10 (0) & 10 (0) & 10 (0) & 10 (0) & 10 (0) & 10 (0) & 10 (0) & 10 (0) & 10 (0) & 10 (0) & 10 (0) & 10 (0) & 10 (0)\\
exp4/1 (10K features) & 15 & 15 (0) & 15 (0) & 15 (0) & 15 (0) & 62 (51) & 15 (0) & 15 (0) & 15 (0) & 15 (0) & 15 (0) & 15 (0) & 15 (0) & 15 (0)\\
exp4/2 (50K features) & 15 & 15 (0) & 15 (0) & 15 (0) & 15 (0) & 364 (320) & 15 (0) & 15 (0) & 15 (0) & 15 (0) & 15 (0) & 15 (0) & 15 (0) & 15 (0)\\
exp4/3 (100K features) & 15 & 15 (0) & 15 (0) & 15 (0) & 15 (0) & 768 (1642) & 16 (2) & 15 (0) & 15 (0) & 15 (0) & 15 (0) & 15 (0) & 16 (2) & 15 (0)\\
exp5/2 ($K=25$) & 6 & 62 (224) & 8 (11) & 6 (0) & 29 (121) & 148 (358) & 27 (139) & 20 (96) & 8 (20) & 6 (1) & 32 (141) & 8 (15) & 26 (134) & 35 (215)\\
exp5/3 ($K=40$) & 6 & 194 (570) & 26 (129) & 8 (5) & 120 (323) & 473 (619) & 105 (319) & 86 (446) & 54 (140) & 14 (17) & 128 (337) & 50 (80) & 98 (290) & 144 (381)\\
exp6/1 (MCAR 15\%) & 4 & 5 (3) & 4 (1) & 220 (329) & 8 (16) & 170 (62) & 22 (48) & 4 (1) & 10 (15) & 4 (1) & 8 (16) & 4 (2) & 32 (52) & 8 (13)\\
exp6/2 (MCAR 30\%) & 4 & 22 (66) & 20 (30) & 650 (906) & 61 (114) & 444 (673) & 115 (162) & 18 (34) & 86 (210) & 6 (11) & 53 (118) & 42 (143) & 116 (168) & 59 (139)\\
\bottomrule
\end{tabular}}
\end{table}

\FloatBarrier
\subsubsection{Regression data-generating processes}

The nine regression settings instantiate four design families.

\paragraph{exp1: heterogeneous nonlinear regression ($N=200$, $p=2000$, $s=4$).}
This design tests recovery of heterogeneous nonlinear effects under Gaussian and heavy-tailed response noise.
With independent Gaussian features,
\[
Y=2X_1+2X_2^2+2\exp(X_3)+2\log|X_4|+4\varepsilon,
\]
where the two variants use Gaussian and $t_3$ errors, respectively.

\paragraph{exp2: ultrahigh-dimensional regression ($N=400$, $s=8$).}
This design tests recovery of linear and quadratic signals as the number of nuisance features increases.
With independent Gaussian features, $Y=2\sum_{j=1}^4X_j+2\sum_{j=5}^8(X_j^2-1)+3\varepsilon$ and $p\in\{10\mathrm K,50\mathrm K,100\mathrm K\}$.

\paragraph{exp3: mixed continuous--categorical regression ($N=400$, $p=2000$, $s=4$).}
This design tests recovery of continuous threshold effects and unordered categorical level effects at different cardinalities.
Two active features are Gaussian, and two are categorical with $K$ levels.  Each categorical level-effect vector $\eta_r$ is an independent permutation of evenly spaced values on $[-1.5,1.5]$, and
\[
Y=5X_1\mathbb I(X_1>1.5)+5X_2+5\{\eta_3[X_3]+\eta_4[X_4]\}+5\varepsilon.
\]
The two variants use $K=5$ and $K=7$, respectively.

\paragraph{exp4: missing feature values under collinearity ($N=300$, $p=3000$, $s=3$).}
This design tests signal recovery under the combined effects of strong collinearity and missing feature values.
Draw $\boldsymbol X\sim\mathcal N(0,\Sigma)$ and set $Y=0.3\{X_1+X_2^2+\exp(X_3)\}+3\varepsilon$. The first variant uses block-AR(1) covariance with $\rho=0.98$ and an MCAR rate of $0.15$; the second uses AR(1) covariance with $\rho=0.95$ and an MCAR rate of $0.20$.
\subsubsection{Full regression results}
\label{app:reg_full}

\begin{table}[!htbp]
\centering
\caption{\textbf{Controlled regression: raw versus PACE screening.} Panels (a) and (b) give matched raw and PACE results, respectively, across exp1--exp4 for each applicable utility in Table~\ref{tab:class_full}. Cells report median MMS (IQR); lower is better.}
\label{tab:reg_full}
\scriptsize
\setlength{\tabcolsep}{4.0pt}
\resizebox{\textwidth}{!}{%
\begin{tabular}{lrrrrrrrr}
\toprule
\multicolumn{9}{l}{(a) Raw feature screening}\\
Setting & $s$ & SIS & DC & HSIC & PC & HD & GCor & IPC\\
\midrule
exp1/2 (Gaussian error) & 4 & 1461 (819) & 54 (122) & 49 (109) & 92 (181) & 149 (396) & 185 (463) & 1482 (854)\\
exp1/3 ($t_3$ error) & 4 & 1442 (828) & 142 (172) & 128 (222) & 180 (212) & 321 (410) & 244 (1224) & 1470 (881)\\
exp2/1 (10K features) & 8 & 8505 (2038) & 26 (39) & 15 (15) & 170 (114) & 56 (80) & 299 (314) & 8993 (2801)\\
exp2/2 (50K features) & 8 & 41714 (11155) & 102 (151) & 26 (140) & 856 (1089) & 226 (434) & 1521 (2037) & 42398 (9969)\\
exp2/3 (100K features) & 8 & 88153 (20315) & 164 (232) & 101 (337) & 2531 (2455) & 1681 (2440) & 4341 (4754) & 85691 (26980)\\
exp3/2 ($K=5$) & 4 & 197 (835) & 40 (211) & 74 (154) & 108 (315) & 134 (309) & 374 (1234) & 836 (1027)\\
exp3/3 ($K=7$) & 4 & 576 (1299) & 94 (288) & 116 (260) & 93 (252) & 96 (325) & 192 (1124) & 750 (1129)\\
exp4/2 (block-AR, MCAR 15\%) & 3 & 18 (30) & 14 (44) & 28 (196) & 24 (110) & 26 (116) & 46 (217) & 56 (248)\\
exp4/3 (AR(1), MCAR 20\%) & 3 & 12 (36) & 12 (60) & 58 (260) & 30 (118) & 40 (292) & 64 (238) & 56 (400)\\
\bottomrule
\end{tabular}}
\vspace{.5em}
\resizebox{\textwidth}{!}{%
\begin{tabular}{lrrrrrrrr}
\toprule
\multicolumn{9}{l}{(b) Matched PACE-$L_2$ feature screening}\\
Setting & $s$ & SIS & DC & HSIC & PC & HD & GCor & IPC\\
\midrule
exp1/2 (Gaussian error) & 4 & 86 (270) & 14 (41) & 27 (70) & 28 (62) & 42 (179) & 40 (52) & 34 (81)\\
exp1/3 ($t_3$ error) & 4 & 226 (379) & 52 (86) & 91 (164) & 56 (158) & 108 (289) & 81 (115) & 74 (168)\\
exp2/1 (10K features) & 8 & 9 (6) & 8 (1) & 12 (41) & 30 (70) & 20 (37) & 25 (34) & 25 (54)\\
exp2/2 (50K features) & 8 & 20 (18) & 11 (7) & 31 (110) & 163 (522) & 97 (403) & 77 (174) & 133 (340)\\
exp2/3 (100K features) & 8 & 19 (20) & 23 (63) & 53 (111) & 97 (156) & 50 (141) & 79 (132) & 101 (156)\\
exp3/2 ($K=5$) & 4 & 6 (26) & 12 (48) & 58 (161) & 108 (371) & 66 (479) & 62 (627) & 86 (390)\\
exp3/3 ($K=7$) & 4 & 10 (28) & 10 (32) & 27 (99) & 96 (258) & 37 (198) & 75 (325) & 98 (271)\\
exp4/2 (block-AR, MCAR 15\%) & 3 & 11 (15) & 12 (31) & 54 (146) & 58 (204) & 34 (250) & 40 (110) & 54 (124)\\
exp4/3 (AR(1), MCAR 20\%) & 3 & 8 (17) & 14 (48) & 74 (190) & 103 (217) & 50 (271) & 46 (191) & 60 (161)\\
\bottomrule
\end{tabular}}
\end{table}

\FloatBarrier

\subsubsection{Utility-specific improvement frequency and paired inference}
\label{app:raw_pace_inference}

The raw-to-PACE effect depends on the screening utility, so Table~\ref{tab:raw_pace_inference} reports each utility separately.  In classification, SIS records 14 wins and two ties; MV, CAVS, CRU, IPC, and LR-FFS each record 13 wins and three ties.  DC records eight wins and eight ties, and HD records six wins and ten ties, so neither utility loses a classification setting.  HSIC produces three wins, eleven ties, and two losses.  In regression, SIS and GCor win all nine settings, while DC and IPC each win eight. Paired tests are performed separately for each utility, using settings as the analysis units.

\begin{table}[!htbp]
\centering
\caption{\textbf{Raw-to-PACE improvement frequency and exploratory paired tests.} A win is lower median MMS under PACE; Wilson intervals count ties as non-wins. Median gain is $\log_2(\mathrm{MMS}_{\rm raw}/\mathrm{MMS}_{\rm PACE})$, so positive values favor PACE. Exact sign tests exclude ties; one-sided sign-flip tests use mean log gain $T$. Tests are unadjusted and use settings as analysis units.}
\label{tab:raw_pace_inference}
\scriptsize
\setlength{\tabcolsep}{4pt}
\resizebox{\linewidth}{!}{%
\begin{tabular}{llrrllll}
\toprule
Task & Utility & Settings & Win/tie/loss & PACE win share [95\% CI] & Median log gain & Sign test (+/nonzero, $p$) & Sign-flip test ($T$, $p$)\\
\midrule
Classification & SIS & 16 & 14/2/0 & 87.5\% [64.0, 96.5] & 7.512 & 14/14, $<0.001$ & $T=6.425$, $<0.001$\\
 & DC & 16 & 8/8/0 & 50.0\% [28.0, 72.0] & 0.161 & 8/8, $0.004$ & $T=0.539$, $0.004$\\
 & FKF & 16 & 11/5/0 & 68.8\% [44.4, 85.8] & 0.776 & 11/11, $<0.001$ & $T=1.069$, $<0.001$\\
 & MV & 16 & 13/3/0 & 81.2\% [57.0, 93.4] & 7.524 & 13/13, $<0.001$ & $T=6.290$, $<0.001$\\
 & PSIS & 16 & 12/4/0 & 75.0\% [50.5, 89.8] & 2.441 & 12/12, $<0.001$ & $T=2.685$, $<0.001$\\
 & CAVS & 16 & 13/3/0 & 81.2\% [57.0, 93.4] & 7.084 & 13/13, $<0.001$ & $T=5.947$, $<0.001$\\
 & HSIC & 16 & 3/11/2 & 18.8\% [6.6, 43.0] & 0.000 & 3/5, $0.500$ & $T=0.198$, $0.188$\\
 & PC & 16 & 9/7/0 & 56.2\% [33.2, 76.9] & 0.959 & 9/9, $0.002$ & $T=1.059$, $0.002$\\
 & HD & 16 & 6/10/0 & 37.5\% [18.5, 61.4] & 0.000 & 6/6, $0.016$ & $T=0.684$, $0.016$\\
 & CRU & 16 & 13/3/0 & 81.2\% [57.0, 93.4] & 7.523 & 13/13, $<0.001$ & $T=6.321$, $<0.001$\\
 & GCor & 16 & 11/5/0 & 68.8\% [44.4, 85.8] & 1.280 & 11/11, $<0.001$ & $T=1.558$, $<0.001$\\
 & LR-FFS & 16 & 13/3/0 & 81.2\% [57.0, 93.4] & 7.086 & 13/13, $<0.001$ & $T=5.932$, $<0.001$\\
 & IPC & 16 & 13/3/0 & 81.2\% [57.0, 93.4] & 7.178 & 13/13, $<0.001$ & $T=6.015$, $<0.001$\\
\midrule
Regression & SIS & 9 & 9/0/0 & 100.0\% [70.1, 100.0] & 5.037 & 9/9, $0.002$ & $T=5.781$, $0.002$\\
 & DC & 9 & 8/0/1 & 88.9\% [56.5, 98.0] & 1.737 & 8/9, $0.020$ & $T=1.790$, $0.006$\\
 & HSIC & 9 & 6/0/3 & 66.7\% [35.4, 87.9] & 0.351 & 6/9, $0.254$ & $T=0.390$, $0.117$\\
 & PC & 9 & 5/1/3 & 55.6\% [26.7, 81.1] & 1.684 & 5/8, $0.363$ & $T=1.100$, $0.078$\\
 & HD & 9 & 7/0/2 & 77.8\% [45.3, 93.7] & 1.376 & 7/9, $0.090$ & $T=1.429$, $0.008$\\
 & GCor & 9 & 9/0/0 & 100.0\% [70.1, 100.0] & 2.209 & 9/9, $0.002$ & $T=2.455$, $0.002$\\
 & IPC & 9 & 8/0/1 & 88.9\% [56.5, 98.0] & 4.312 & 8/9, $0.020$ & $T=4.718$, $0.006$\\
\bottomrule
\end{tabular}}
\end{table}

\FloatBarrier

\subsection{Comparison with statistical and machine-learning baselines}
\label{app:cross_paradigm}

\subsubsection{Implementation of the comparison methods}

The methods below are evaluated in the controlled comparisons or the TALENT study (Table~\ref{tab:talent_full_field}). Within each controlled replication, the compared methods receive the same generated dataset.

\paragraph{InterDependence Scores.}
IDS~\citep{radhakrishnan2025efficiently} expands each feature and the response into six Gaussian-damped polynomial coordinates and takes the largest absolute cross-correlation:
\begin{equation}
\label{eq:utility_ids}
\phi_a(u)=\exp(-0.5u^2)\frac{u^a}{\sqrt{a!}},
\qquad
\mathcal U_{\rm IDS}(X_j,Y)=
\max_{\substack{0\le a,b\le5\\1\le c\le q_Y}}
\left|\widehat\rho\bigl(\phi_a(X_j),\phi_b(V^{(:,c)})\bigr)\right|.
\end{equation}
Here $\widehat\rho$ and $V$ follow Appendix~\ref{app:utility_formulas}, with $q_Y=1$ response coordinate for regression and $q_Y=R$ for classification. Each expansion is applied elementwise, and the correlation standardizes its resulting coordinates. Extending the polynomial order beyond 8 yields near-zero terms, so we keep the same setting as the original IDS method.

\paragraph{Neighborhood and fixed-representation comparators.}
ReliefF and RReliefF~\citep{RobnikSikonja1997AnAO} use at most 1,000 reference observations and ten nearest neighbors. Continuous-feature distances are normalized by their training ranges, categorical mismatches contribute unit distance, and the regression variant weights neighbors by rank and response difference. The prespecified representation families comprise degree-8 Legendre polynomials, cubic B-splines, quantile-centered RBFs, 128 random Fourier features~\citep{rahimi2007random}, and 16-bin quantile indicators. Representations are padded with zeros or truncated to 128 coordinates and paired with the dependence score named in each method label; Legendre-WMC and Spline-WMC use ridge-whitened multiple correlation. UniTree-CV ranks a feature by the five-fold loss reduction of a depth-3 univariate decision tree relative to an intercept-only predictor, with a minimum leaf size of $\max(5,\lceil0.02N\rceil)$.

For continuous columns, RFF uses $\sqrt{2/d}\cos(\omega_m x+b_m)$ with $d=128$, Gaussian frequencies $\omega_m\sim\mathcal N(0,\sigma^{-2})$, and phases $b_m\sim\operatorname{Unif}(0,2\pi)$. The bandwidth $\sigma$ is the median pairwise absolute distance within the training column, with the column's standard deviation and then one as fallbacks if this median is zero; output coordinates are centered. Categorical columns use one-hot indicators, merging levels with fewer than $\max(2,\lceil0.005N\rceil)$ observations before dimension alignment. Thus the control matches the output dimension but uses a different map from the Transformer; its bandwidth and categorical vocabulary depend on feature values, not responses.

\paragraph{Fitted-model importance.}
Random forests~\citep{breiman2001random}, LightGBM~\citep{ke2017lightgbm}, and CatBoost~\citep{prokhorenkova2018catboost} use 300 trees or boosting iterations and a maximum depth of 6. Random forests use bootstrap sampling and a feature fraction of $0.8$; LightGBM uses a learning rate of $0.05$ and row and feature fractions of $0.8$; CatBoost uses a learning rate of $0.05$ and a feature fraction of $0.8$. The reported rankings use random-forest mean decrease in impurity, LightGBM split count or total gain, and CatBoost PredictionValuesChange, respectively.

\paragraph{Predictive attribution.}
LightGBM- and CatBoost-SHAP~\citep{lundberg2017unified} rank features by their mean absolute TreeSHAP values on at most 1,000 validation observations. TabICL-SHAP fits two estimators, uses a single all-missing row as the masking background, evaluates at most 256 validation observations, and averages absolute permutation-SHAP values across observations and classes. The TALENT comparison additionally includes LOCO~\citep{lei2018distribution}, which uses the increase in validation loss after refitting a random forest without each feature; LOCO-CV averages the same loss increase over five folds. SAGE~\citep{covert2020understanding}, also evaluated on TALENT, uses 32 feature permutations and up to 32 marginal imputations from a training-fold background for fitted LightGBM or CatBoost predictors.

\paragraph{Supervised learned selectors.}
xRFM-AGOP~\citep{beaglehole2026xrfm} fits one hard-routing tree with a maximum leaf size of 60,000 and ranks features by the diagonal of the training-fold average gradient outer product.  LassoNet~\citep{JMLR:v22:20-848} uses one 64-unit hidden layer, hierarchy parameter $M=10$, 300 dense epochs, 30 path epochs with early stopping, and the official regularization-path importance.  GradEnFS~\citep{liu2024supervised} is classification-only and trains a 512-unit sparse hidden layer for 50 epochs; its accumulated input-gradient scores define the ranking.  SAND~\citep{pad2025sand} trains for 100 epochs with noise scale $0.5$ and gate budget $k_0=100$, then ranks features by their projected gate values.  Deep Lasso~\citep{cherepanova2023performance} uses two 128-unit hidden layers, dropout $0.1$, gradient regularization $0.1$, and 100 epochs; its validation-fold input-gradient importance defines the ranking.

\subsubsection{Setting-level cross-paradigm method landscapes}
\label{app:simulation_landscapes}

The setting-level landscapes complement the matched raw/PACE analysis by comparing method families within the controlled simulation settings. The primary raw-to-PACE results are reported in Tables~\ref{tab:class_full}--\ref{tab:raw_pace_inference}.

\begin{figure}[!htbp]
\centering
\includegraphics[width=.98\textwidth,height=.82\textheight,keepaspectratio]{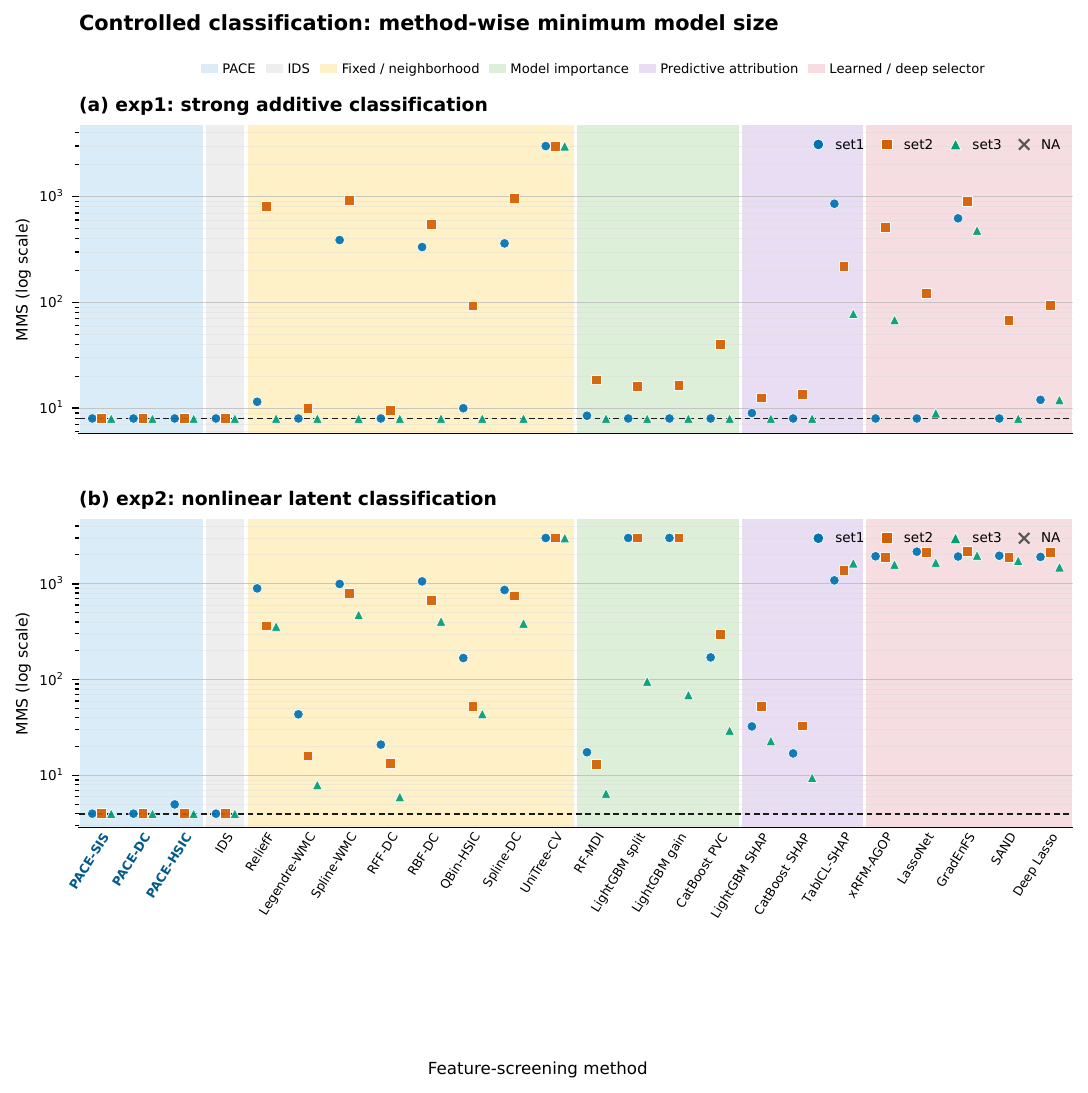}
\caption{\textbf{Controlled classification exp1--exp2: cross-paradigm MMS.} Median MMS by method on a log scale; lower is better. Dashed lines mark oracle support, colors/markers denote parameter sets, shading denotes method families, and top crosses mark missing results. (a) exp1: Gaussian, $t_3$, log-normal. (b) exp2: independent, AR(1)/$t_3$, banded.}
\label{fig:sim_landscape_cls_exp1_exp2}
\end{figure}

\begin{figure}[!htbp]
\centering
\includegraphics[width=.98\textwidth,height=.82\textheight,keepaspectratio]{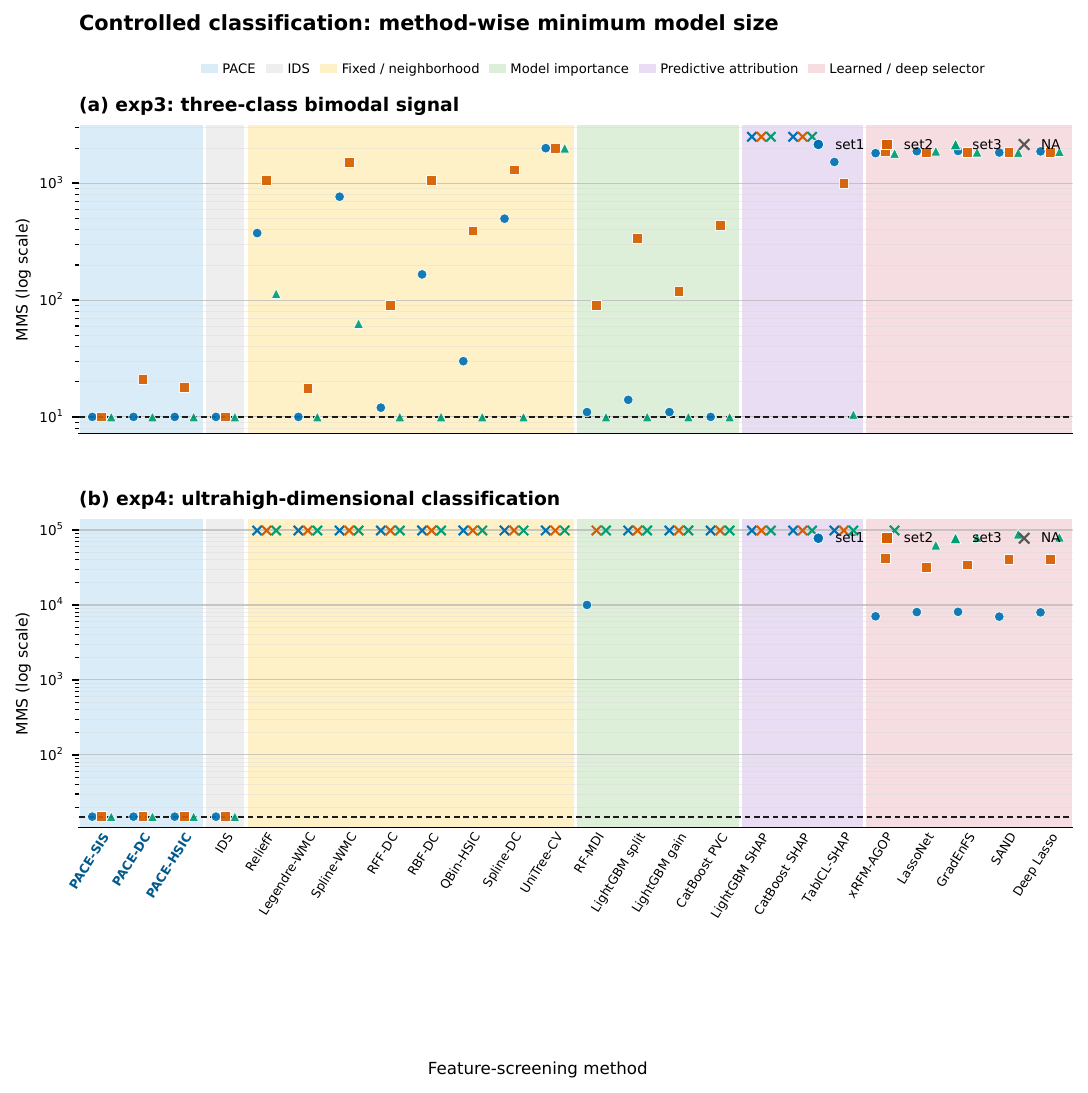}
\caption{\textbf{Controlled classification exp3--exp4: cross-paradigm MMS.} Median MMS by method on a log scale; lower is better. Dashed lines mark oracle support, colors/markers denote parameter sets, shading denotes method families, and top crosses mark missing results. (a) exp3: moderate imbalance, severe imbalance, balanced. (b) exp4: 10K, 50K, 100K features.}
\label{fig:sim_landscape_cls_exp3_exp4}
\end{figure}

\begin{figure}[!htbp]
\centering
\includegraphics[width=.98\textwidth,height=.82\textheight,keepaspectratio]{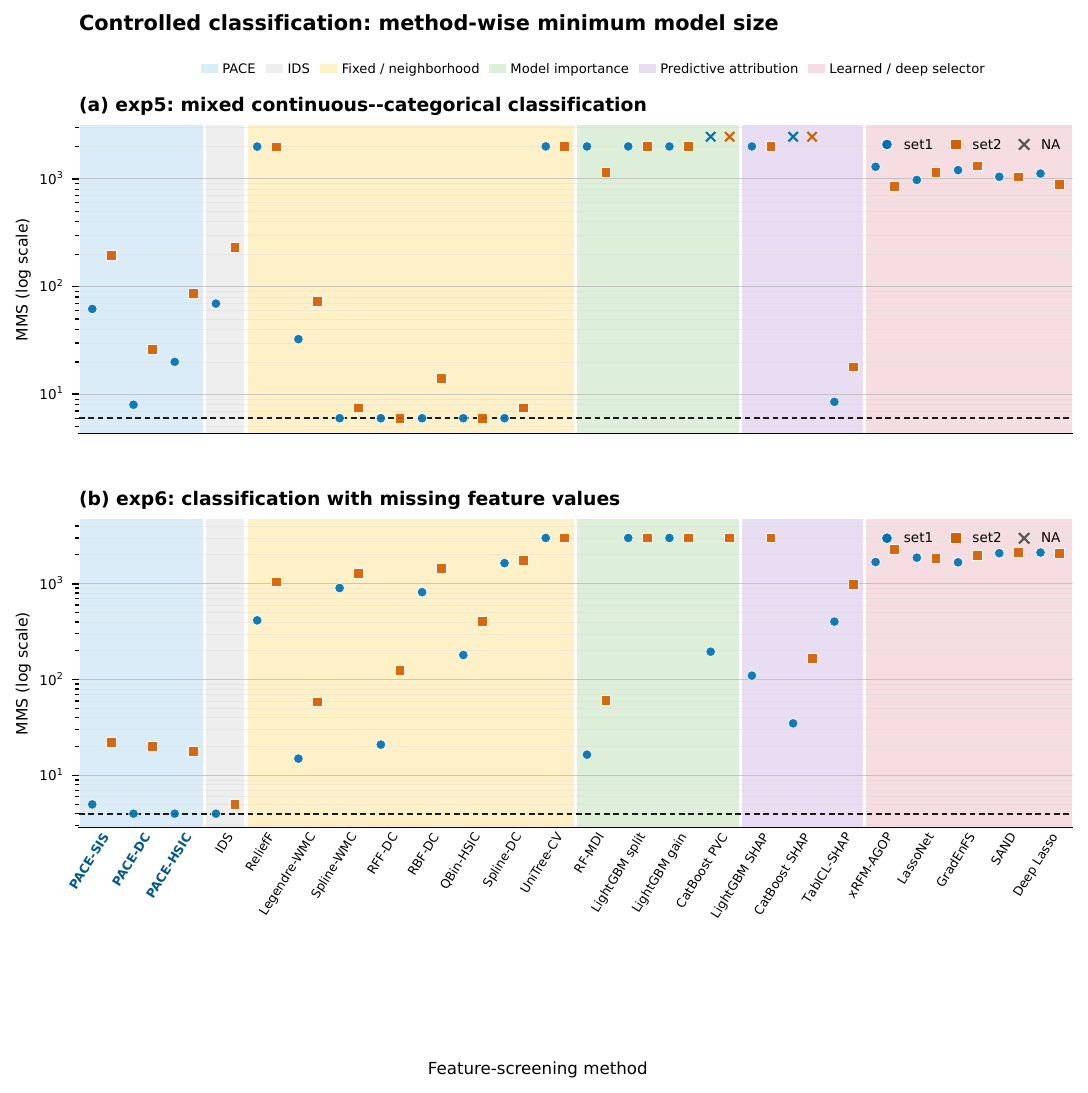}
\caption{\textbf{Controlled classification exp5--exp6: cross-paradigm MMS.} Median MMS by method on a log scale; lower is better. Dashed lines mark oracle support, colors/markers denote parameter sets, shading denotes method families, and top crosses mark missing results. (a) exp5: $K=25,40$. (b) exp6: MCAR 15\%, 30\%.}
\label{fig:sim_landscape_cls_exp5_exp6}
\end{figure}

\begin{figure}[!htbp]
\centering
\includegraphics[width=.98\textwidth,height=.82\textheight,keepaspectratio]{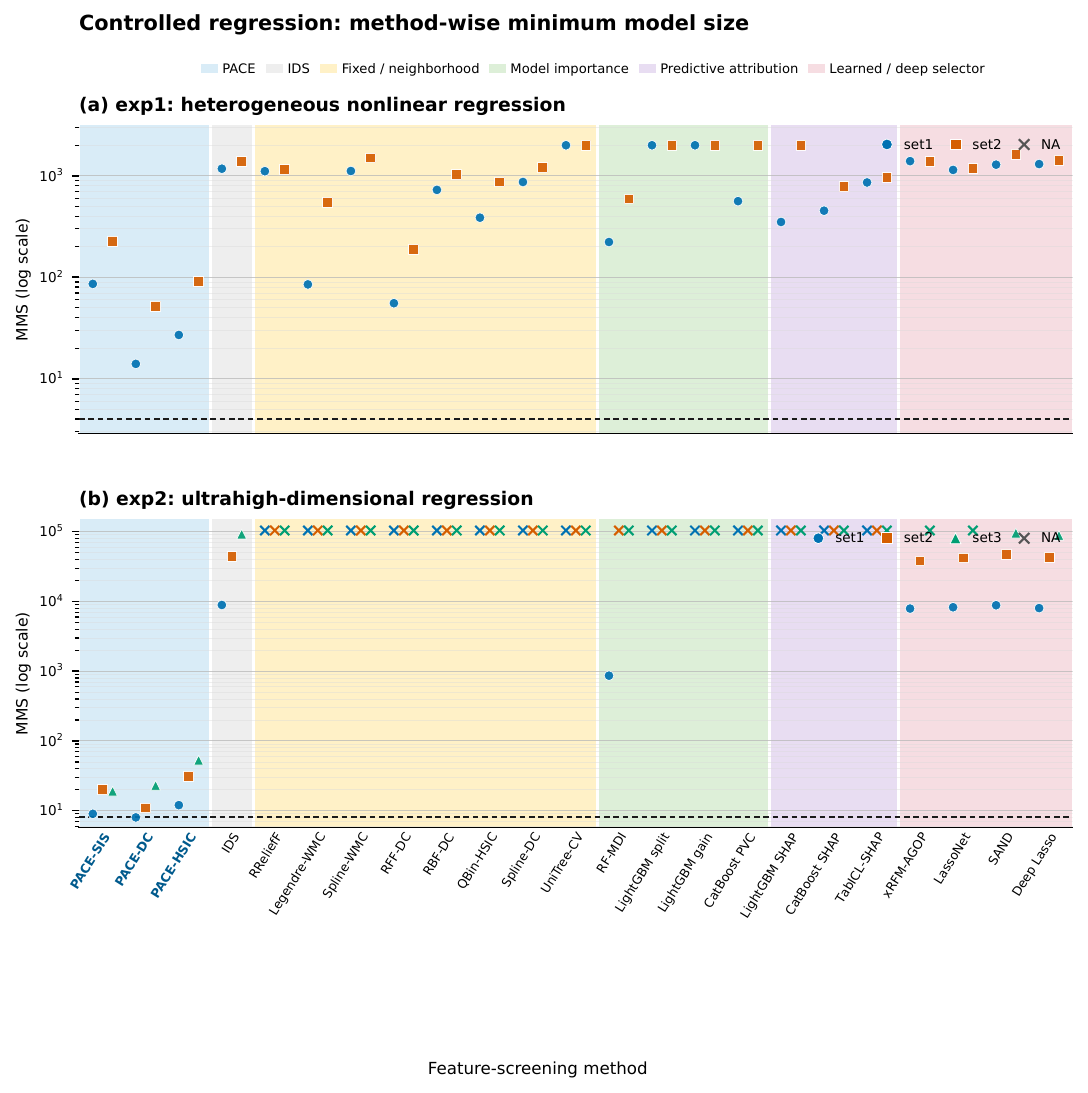}
\caption{\textbf{Controlled regression exp1--exp2: cross-paradigm MMS.} Median MMS by method on a log scale; lower is better. Dashed lines mark oracle support, colors/markers denote parameter sets, shading denotes method families, and top crosses mark missing results. (a) exp1: Gaussian/$t_3$ error. (b) exp2: 10K, 50K, 100K features.}
\label{fig:sim_landscape_reg_exp1_exp2}
\end{figure}

\begin{figure}[!htbp]
\centering
\includegraphics[width=.98\textwidth,height=.82\textheight,keepaspectratio]{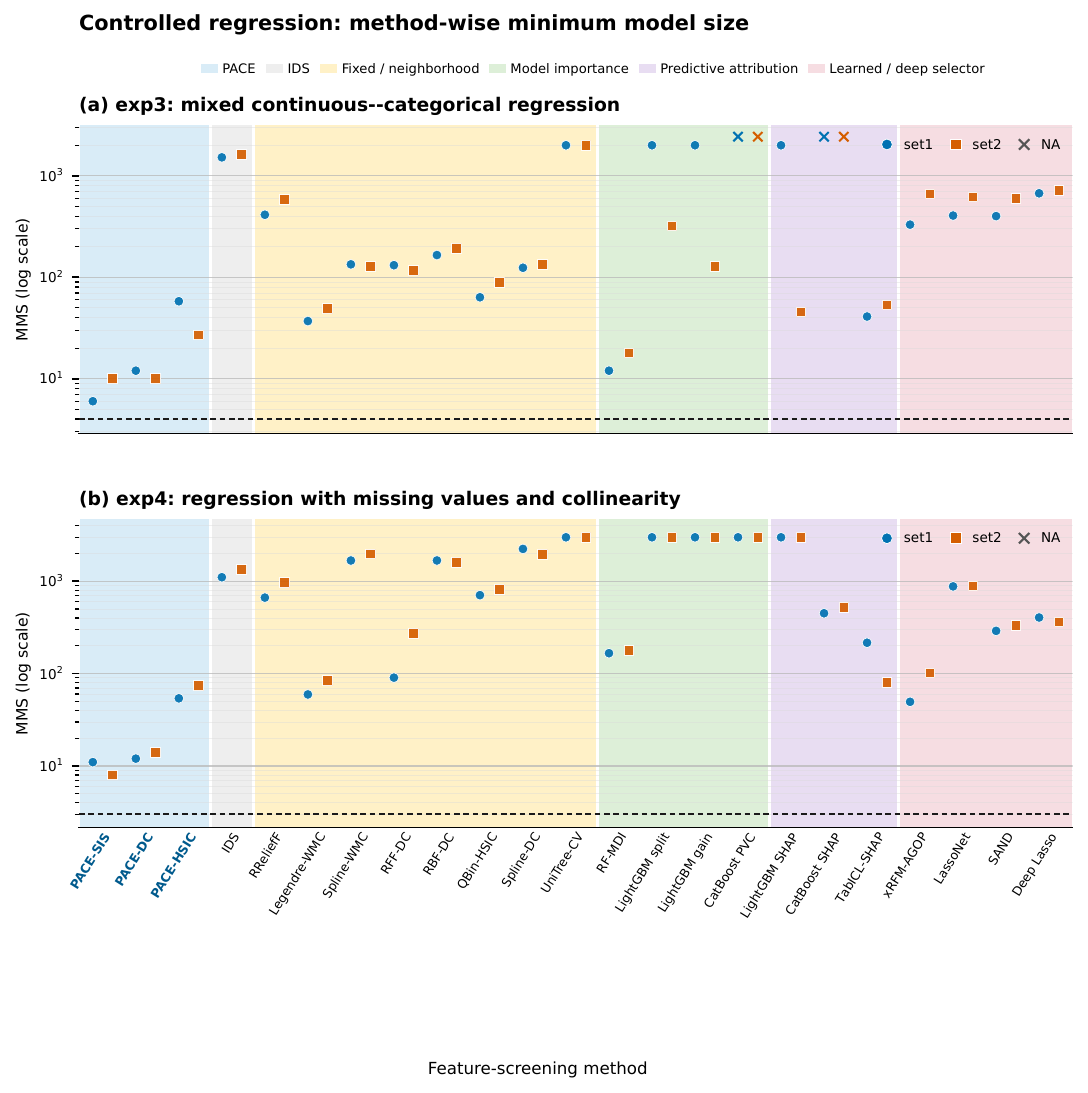}
\caption{\textbf{Controlled regression exp3--exp4: cross-paradigm MMS.} Median MMS by method on a log scale; lower is better. Dashed lines mark oracle support, colors/markers denote parameter sets, shading denotes method families, and top crosses mark missing results. (a) exp3: $K=5,7$. (b) exp4: block-AR/MCAR 15\%, AR(1)/MCAR 20\%.}
\label{fig:sim_landscape_reg_exp3_exp4}
\end{figure}

Figures~\ref{fig:sim_landscape_cls_exp1_exp2}--\ref{fig:sim_landscape_reg_exp3_exp4} place methods on the horizontal axis and MMS on a logarithmic vertical axis. Each figure contains two experiment families, with up to three data-generating parameter settings per family. The plots include IDS as a standalone fixed-expansion comparator and compare PACE with fixed or neighborhood representations, fitted-model importance, predictive attribution, and learned or deep selectors. Crosses at the top of a panel indicate that no finite result is available.

Figures~\ref{fig:sim_landscape_cls_exp1_exp2} and~\ref{fig:sim_landscape_cls_exp3_exp4} show the clearest separation in nonlinear and ultrahigh-dimensional classification: the three PACE variants have lower MMS than many fitted and learned selectors, with oracle MMS in several nonlinear and ultrahigh-dimensional settings.  Figure~\ref{fig:sim_landscape_cls_exp5_exp6} shows greater variation in mixed-type and missing-value classification.  Figures~\ref{fig:sim_landscape_reg_exp1_exp2} and~\ref{fig:sim_landscape_reg_exp3_exp4} show that PACE-SIS and PACE-DC often rank the active set early relative to ambient width, with notable gaps from the oracle in nonlinear and missing-value regression; several fitted alternatives either rank the active set later or return no finite result.

\FloatBarrier
\FloatBarrier
\FloatBarrier
\FloatBarrier

\FloatBarrier
\begingroup

\subsection{Recovery under product interactions}
\label{app:interaction_simulations}

We evaluate three regression designs with product interactions, each using $N=512$ observations and 80 paired repetitions. The first two use $p=100$ independent standard normal features with $Y=X_1X_2+\epsilon$ and $Y=1.5X_1X_2X_3+\epsilon$. The third uses $p=500$ features with independent standard normal logarithms and $Y=1.5(\log X_1)(\log X_2)(\log X_3)+\epsilon$. In all designs, $\epsilon\sim\mathcal N(0,1)$ independently of the features.

\begin{table}[!htbp]
\centering

\caption{\textbf{Active-feature recovery under product interactions.} Entries count repetitions in which all active features rank among the top ten, out of 80 paired repetitions. Column headings specify the signal before independent noise is added; $L=(\log X_1)(\log X_2)(\log X_3)$.}
\label{tab:interaction_recovery}
\small
\begin{tabular}{lccc}
\toprule
Method & $X_1X_2$ & $1.5X_1X_2X_3$ & $1.5L$\\
\midrule
Raw-SIS & 8 & 3 & 0\\
PACE-SIS & 9 & 6 & 0\\
Raw-DC & 69 & 67 & 39\\
PACE-DC & \textbf{80} & \textbf{79} & 72\\
Raw-HSIC & \textbf{80} & \textbf{79} & 59\\
PACE-HSIC & \textbf{80} & \textbf{79} & \textbf{77}\\
LightGBM-Gain & 79 & 69 & 27\\
XGBoost-Gain & 4 & 1 & 0\\
\bottomrule
\end{tabular}
\end{table}

Table~\ref{tab:interaction_recovery} shows that PACE improves DC recovery in all three designs and HSIC recovery in the log-product design. These interactions have zero conditional mean, $\mathbb E[Y\mid X_j]=0$, for each active feature, but their conditional variance depends on that feature. DC and HSIC can detect this distributional dependence, whereas the zero conditional mean limits correlation-based SIS. The results extend the recovery gains beyond additive signals and illustrate the importance of matching the utility to the form of dependence.
\endgroup
\FloatBarrier

\section{Formal Information Boundary}
\label{app:formal}

\subsection{Proof of Proposition~\ref{prop:no_creation}}

Using the proposition's column notation, let $B$ and $C$ be measurable sets in the representation and response-vector spaces, respectively. Measurability of $G_N$ and $X_j\perp Y$ give
\begin{align*}
P\{G_N(X_j)\in B,Y\in C\}
&=P\{X_j\in G_N^{-1}(B),Y\in C\}\\
&=P\{X_j\in G_N^{-1}(B)\}P\{Y\in C\},
\end{align*}
which proves $G_N(X_j)\perp Y$.

For i.i.d.\ observations, marginal feature--response independence yields the column-level condition through
\[
P_{X_j,Y}
=\prod_{i=1}^N P_{X_{ij},Y_i}
=\left(\prod_{i=1}^N P_{X_{ij}}\right)
 \left(\prod_{i=1}^N P_{Y_i}\right)
=P_{X_j}P_Y.
\]
If extraction uses randomness $R\perp(X_j,Y)$, the same independence argument gives $G_N(X_j,R)\perp Y$. For the injectivity statement, $R$ must be fixed or observed and conditioned upon. Injectivity conditional on an unobserved random seed does not imply marginal information preservation.

If $G_N$ is almost surely injective between standard Borel spaces, its inverse on the image is measurable up to a null set. Therefore $X_j$ is measurable with respect to $G_N(X_j)$, and the reverse inclusion follows from measurability of $G_N$; thus $\sigma(G_N(X_j))=\sigma(X_j)$ up to null sets. This gives a sufficient condition for information preservation.

\subsection{Scope of Proposition~\ref{prop:no_creation}}

The proposition establishes that a fixed measurable column operator preserves independence between a feature column and the response vector. Finite-sample rankings can still select inactive features by chance. The empirical study evaluates finite-sample ordering of target and inactive features in the transformed geometry.

\input{appendix_real}

\FloatBarrier
\section{Representation Ablations: Backbones, Layers, and Utilities}
\label{app:ablation_full}

Main-text Figure~\ref{fig:ablation_summary} examines backbone, pretraining, and extraction effects. Panel (a) uses comparisons with raw input and random-weight controls to assess whether dimensional expansion alone is sufficient. Panel (b) compares PACE with a fixed RFF map under matched utilities and reports the ratio of their TALENT runtimes. Panel (c) assesses whether executing the preceding blocks changes performance relative to executing the same block in isolation.

\FloatBarrier
\subsection{Pretrained representations versus raw input and matched random-weight controls}

\begin{figure}[!htbp]
\centering
\includegraphics[width=.9\linewidth]{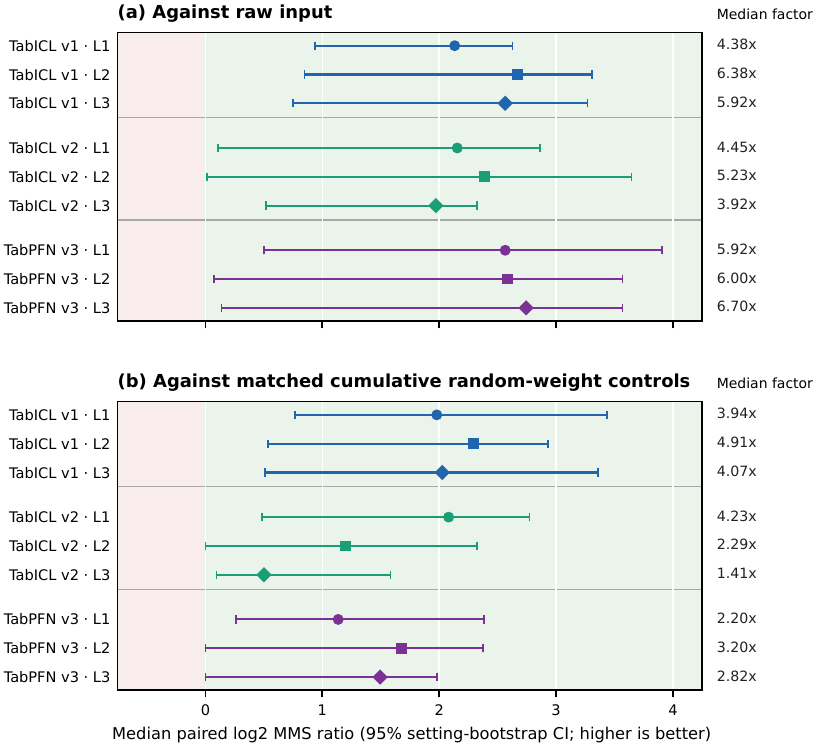}
\caption{\textbf{Pretraining versus raw and random-weight representations.} Points are medians across settings of within-setting median paired log-MMS ratios; a gain of one on the $\log_2$ scale corresponds to a twofold MMS reduction. Exponentiating the TabICL v1-$L_2$ estimates gives MMS reduction factors of $6.38$ relative to raw input (top) and $4.91$ relative to its matched cumulative random-weight control (bottom). Bars are descriptive 95\% setting-bootstrap intervals.}
\label{fig:mechanism_main}
\end{figure}

\FloatBarrier
\subsection{Backbone--layer comparisons}

\begin{figure}[!htbp]
\centering
\includegraphics[width=.9\linewidth]{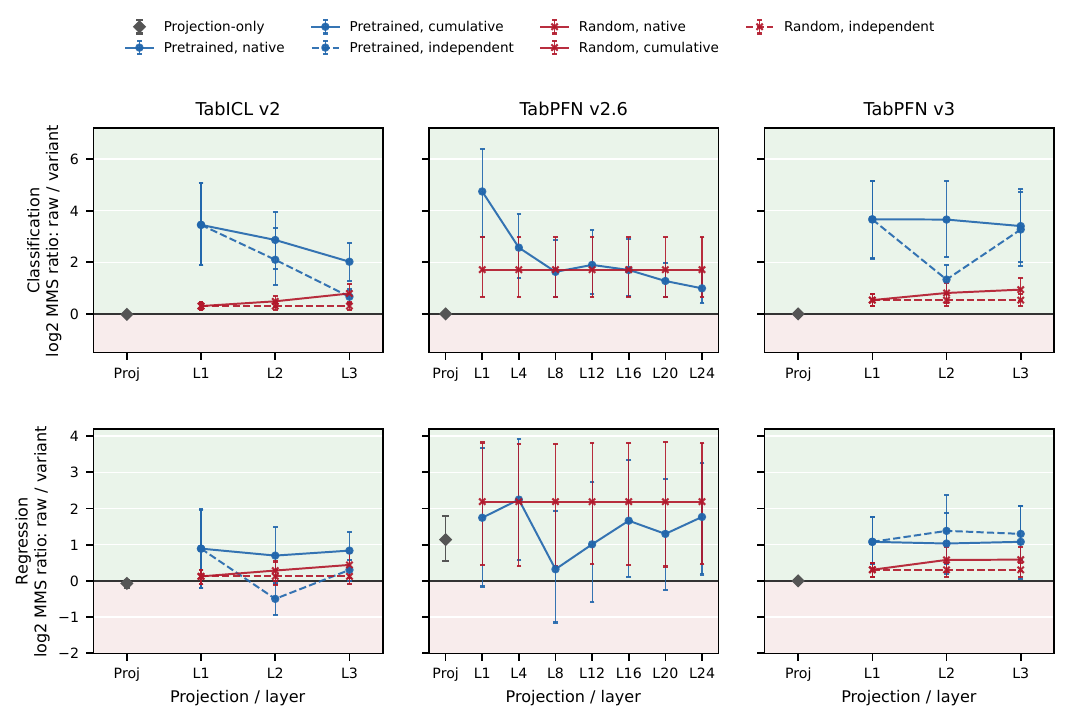}
\caption{\textbf{Backbone and layer effects.}  A positive value of $\log_2(\mathrm{MMS}_{\rm raw}/\mathrm{MMS}_{\rm variant})$ favors the representation.  Points show utility-level log-ratios averaged within each setting and then across settings within each task.  Projection-only, pretrained, and matched random-weight variants are shown with descriptive 95\% setting-bootstrap intervals.}
\label{fig:ablation_all}
\end{figure}

\FloatBarrier
\newpage
\subsection{Transfer across screening utilities}

\begin{figure}[!htbp]
\centering
\includegraphics[width=.9\linewidth]{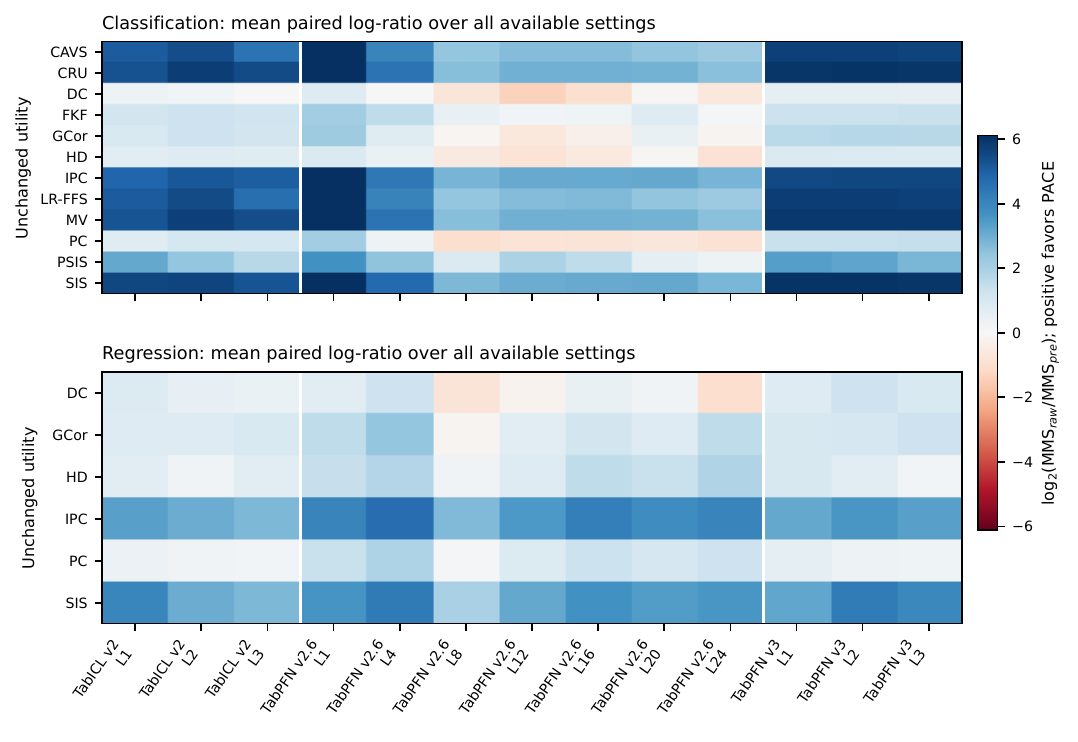}
\caption{\textbf{Transfer across utilities in the controlled ablation setting.} Cells show mean paired $\log_2(\mathrm{MMS}_{\rm raw}/\mathrm{MMS}_{\rm PACE})$; blue favors PACE. Classification: CAVS, CRU, DC, FKF, GCor, HD, IPC, LR-FFS, MV, PC, PSIS, SIS. Regression: DC, GCor, HD, IPC, PC, SIS. }
\label{fig:ablation_utility}
\end{figure}

\FloatBarrier
\newpage
\subsection{Mean versus maximum coordinate aggregation}
\label{app:sis_aggregation}

Maximum aggregation uses the embedding coordinate with the strongest empirical association with the response. This can prevent a signal concentrated in a few coordinates from being diluted by averaging, strengthening PACE-SIS in several nonlinear classification settings. The maximum also gives greater weight to large associations caused by noise, increasing sensitivity to isolated extreme scores. Using the notation of Equation~\eqref{eq:utility_dispatch}, we compare
$$
\omega_j^{\rm mean}=\frac{1}{d}\sum_{m=1}^{d}\mathcal U_{\rm SIS}\!\left(\mathbf H_j^{(:,m)},Y\right),
\qquad
\omega_j^{\rm max}=\max_{1\leq m\leq d}\mathcal U_{\rm SIS}\!\left(\mathbf H_j^{(:,m)},Y\right).$$

\begin{table}[!htbp]
\centering
\caption{\textbf{Coordinate aggregation on TALENT.} Mean and Max are across-dataset means after averaging three paired splits. Classification reports AUC ($\uparrow$); regression reports $R=\mathrm{RMSE}/\mathrm{RMSE}^{\rm clean}$ for the same learner and split ($\downarrow$), where $R=1$ matches clean-table performance. Retention is the percentage of original features retained ($\uparrow$). The paired $t$ statistic tests Max minus Mean, using datasets as analysis units.}
\label{tab:sis_aggregation_talent}
\footnotesize
\setlength{\tabcolsep}{3pt}
\begin{tabular}{lrrrrrrrrr}
\toprule
& \multicolumn{3}{c}{Binary AUC $\uparrow$} & \multicolumn{3}{c}{Multiclass AUC $\uparrow$} & \multicolumn{3}{c}{Regression $R$ $\downarrow$}\\
\cmidrule(lr){2-4}\cmidrule(lr){5-7}\cmidrule(lr){8-10}
Learner & Mean & Max & $t$ & Mean & Max & $t$ & Mean & Max & $t$\\
\midrule
LightGBM & 0.806 & 0.807 & +0.20 & 0.891 & 0.890 & -0.29 & 1.282 & 1.975 & +1.39 \\
XGBoost & 0.802 & 0.803 & +0.83 & 0.896 & 0.896 & +0.99 & 1.281 & 1.978 & +1.36 \\
CatBoost & 0.814 & 0.815 & +0.24 & 0.898 & 0.897 & -1.46 & 1.303 & 2.000 & +1.37 \\
MLP & 0.764 & 0.769 & +1.73 & 0.844 & 0.852 & +3.30 & 1.359 & 1.745 & +1.23 \\
RealMLP & 0.778 & 0.774 & -2.11 & 0.882 & 0.883 & +0.56 & 1.333 & 2.072 & +1.21 \\
TabM & 0.796 & 0.798 & +1.04 & 0.880 & 0.884 & +0.93 & 1.109 & 1.333 & +1.06 \\
TabPFN v3 & 0.832 & 0.835 & +1.50 & 0.925 & 0.925 & +0.19 & 1.492 & 3.046 & +1.22 \\
TabPFN v2.6 & 0.833 & 0.834 & +0.04 & 0.924 & 0.924 & +0.67 & 1.335 & 2.505 & +1.28 \\
TabICL v2 & 0.831 & 0.834 & +0.98 & 0.923 & 0.924 & +0.95 & 1.739 & 5.142 & +1.10 \\
TabDPT & 0.804 & 0.809 & +2.16 & 0.899 & 0.898 & -0.34 & 1.699 & 3.392 & +1.17 \\
\midrule
Retention (\%) $\uparrow$ & 69.40 & 66.46 & -2.69 & 83.47 & 80.28 & -4.07 & 60.43 & 56.40 & -2.37 \\
\bottomrule
\end{tabular}
\end{table}

\paragraph{Classification and retention.}
Max raises mean binary AUC for nine of ten learners, with the largest gains for MLP and TabDPT ($0.005$ AUC). Multiclass MLP improves from $0.844$ to $0.852$; the four TFMs change by less than $0.001$ AUC each. Original-feature retention decreases by $2.94$, $3.19$, and $4.04$ percentage points in binary, multiclass, and regression tasks, respectively.

\paragraph{Regression stability.}
Mean aggregation limits large regression losses. Max increases mean clean-normalized RMSE for every learner; for TabICL v2, mean $R$ rises from $1.739$ to $5.142$. The largest losses are concentrated in a small subset of datasets.

\paragraph{Controlled support recovery.}
The simulation ablation uses a fresh set of paired replicates, evaluating both aggregations on the same generated data and embeddings. Table~\ref{tab:sis_aggregation_simulation} reports median MMS (IQR).

\begin{table}[!htbp]
\centering
\caption{\textbf{Mean versus maximum aggregation in controlled simulations.} Entries report median MMS (IQR); lower is better. Experiment and setting identifiers follow Appendix~\ref{app:controlled}.}
\label{tab:sis_aggregation_simulation}
\footnotesize
\renewcommand{\arraystretch}{0.95}
\setlength{\tabcolsep}{6pt}
\begin{tabular}{lrr}
\toprule
Exp./setting & Mean MMS & Max MMS \\
\midrule
\multicolumn{3}{l}{\textit{Classification}}\\
exp1/1 & 8 (0) & 8 (0) \\
exp1/2 & 8 (0) & 8 (0) \\
exp1/3 & 8 (0) & 8 (0) \\
exp2/1 & 5 (2) & 5 (9.75) \\
exp2/2 & 5 (2) & 4 (0) \\
exp2/3 & 4 (0) & 4 (0) \\
exp3/1 & 10 (0) & 10 (0) \\
exp3/2 & 10 (1) & 10 (0) \\
exp3/3 & 10 (0) & 10 (0) \\
exp4/1 & 15 (0) & 15 (0) \\
exp4/2 & 15 (0) & 15 (0) \\
exp4/3 & 15 (0) & 15 (0) \\
exp5/2 & 52.5 (323.5) & 7 (26.75) \\
exp5/3 & 115.5 (359.25) & 13 (66.75) \\
exp6/1 & 5 (2.75) & 4 (0) \\
exp6/2 & 27 (37) & 4 (3) \\
\addlinespace
\multicolumn{3}{l}{\textit{Regression}}\\
exp1/2 & 80.5 (127) & 15.5 (71.25) \\
exp1/3 & 161.5 (441.5) & 93 (361) \\
exp2/1 & 9 (3) & 9 (4) \\
exp2/2 & 13 (12) & 16 (37) \\
exp2/3 & 26 (64) & 53 (505) \\
exp3/2 & 8 (17.25) & 4.5 (5) \\
exp3/3 & 9.5 (24.25) & 4 (8.75) \\
exp4/2 & 9 (13) & 8 (8) \\
exp4/3 & 7 (25) & 6 (14.25) \\
\bottomrule
\end{tabular}
\end{table}

The two aggregation strategies yield broadly similar support-recovery performance across most settings. The largest classification gains from maximum aggregation occur with mixed feature types and missingness: median MMS decreases from $52.5$ to $7$ and from $115.5$ to $13$ in exp5, and from $27$ to $4$ at 30\% missingness in exp6, relative to mean aggregation. Aggregation also affects variability: in classification exp2/1, both strategies achieve median MMS $5$, while the IQR increases from $2$ under mean aggregation to $9.75$ under maximum aggregation.

Mean remains the default aggregation. Max improves support recovery in mixed-type and incomplete designs, while mean reduces the influence of isolated extreme scores and provides more stable performance on TALENT.

\section{Reproducibility}
\label{app:repro}

\paragraph{Implementation details.}
PACE returns rankings over original input-column indices for evaluation by a prespecified downstream learner. Encoder configuration, preprocessing, screening budget, repeated splits, and downstream configuration are fixed within each matched raw/PACE comparison. TALENT uses three random seeds, and the natural high-dimensional experiments use five. If a TALENT training partition exceeds 2,000 rows, embedding and utility estimation use at most 2,000 rows selected by stratified sampling for classification and by sampling without replacement for regression, using the corresponding random seed in both cases. TabPFN-Wide and GOTabPFN are evaluated as end-to-end prediction procedures. Timing results throughout the paper and appendix were recorded on an NVIDIA A100, except for the feature-dimension scaling experiment, which used an NVIDIA GeForce RTX 4080 SUPER shared by all compared methods.

\paragraph{Preprocessing and information access.}
Utility estimation and feature ranking use training responses; representation extraction uses feature values only. {For TabICL v1, we extract from the column encoder before row interaction. For TabICL v2 and TabPFN v3, we bypass the step that adds a training-label embedding to the feature-group embedding. We pass the feature projection directly to the TabICL column Transformer or the TabPFN feature-distribution embedding blocks. Each group contains copies of a single original feature. For TabPFN v2.6, the projection-only ablation processes each feature separately. The layer ablation supplies a constant zero vector to the target encoder, skips cross-group attention, and reads out only the feature tokens. Each feature in a native group receives the same group representation.} 

The TALENT vocabulary is constructed from categorical feature levels pooled across the training, validation, and test partitions, yielding transductive, label-free preprocessing. {For the raw screening methods, categorical features were one-hot encoded, missing numerical values were imputed using feature-wise means, and continuous features were retained on their original scales. Scale differences were addressed through method-specific normalization or kernel bandwidth selection within the screening utilities.}

\paragraph{Data availability.}
The study uses synthetic data specified in this paper and public datasets from the TALENT collection, subject to the original access conditions and licenses.

\paragraph{Ethics and human subjects.}
The study uses synthetic and previously released tabular datasets, with no new participant recruitment or collection of personal data. Dataset-specific licenses and use restrictions apply.

\section{SelectArena Evaluation Protocol}
\label{app:selectarena_protocol}
{We evaluate 49 binary classification, 12 multiclass classification, and 10 regression datasets from SelectArena~\citep{schafer2026selectarena}. Following the original benchmark protocol, each dataset is evaluated using five dataset-specific feature budgets. Let $p$ denote the number of original features and $j\in\{1,\ldots,5\}$ index the budget. The number of selected features is defined as
\begin{equation}
k_j =
\begin{cases}
\lceil pj/6\rceil, & p\leq 100,\\
\lceil p^{j/6}\rceil, & p>100.
\end{cases}
\label{eq:selectarena_budgets}
\end{equation}
This design uses approximately linear budget spacing for datasets with at most 100 features and exponential spacing for higher-dimensional datasets, placing greater emphasis on smaller feature subsets~\citep{schafer2026selectarena}.}

{For each budget, selectors use the training fold to rank the original features and retain the top-$k_j$ features. For our raw and PACE methods, screening utilities are computed using at most 2,000 training observations per fold. The selected subsets are then evaluated using four downstream predictors: TabICL v2, random forest (RF), LightGBM (LGBM), and a linear model, over three matched outer folds.}

{Test error is $1-\mathrm{AUC}$ for binary classification, log-loss for multiclass classification, and RMSE for regression. We report the percentage error reduction relative to random selection, defined as
\begin{equation}
\mathrm{ER}(m,d)
=
100\cdot
\frac{\mathrm{Err}(\mathrm{Random},d)-\mathrm{Err}(m,d)}
{\mathrm{Err}(\mathrm{Random},d)},
\end{equation}
where $m$ denotes a feature selection method and $d$ denotes a dataset. For Figure~\ref{fig:selectarena_comparison}(a), we first average errors over matched budgets, folds, and predictors within each dataset, compute the percentage reduction relative to random selection, and average these reductions equally across datasets. Table~\ref{tab:selectarena_class_budgets} applies the same procedure separately at each budget on the 61 classification datasets; its final column averages the five budget-specific reductions equally.}

{Elo ratings are obtained by fitting a Bradley--Terry model to pairwise outcomes after averaging errors across matched folds for each dataset, predictor, and budget. Random selection is anchored at 1,000, and 95\% confidence intervals are obtained using 1,000 dataset-level bootstrap resamples, preserving matched method comparisons within each resampled dataset.}

\subsection{Paired PACE--Raw Comparisons}
\label{app:selectarena_significance}
{We further assess the improvements of PACE over raw screening through paired comparisons across datasets. For each utility, we compute the PACE-minus-Raw difference in error reduction relative to random selection, using one paired observation per dataset averaged over matched predictors, folds, and budgets. We consider both the full benchmark and the classification subset. Mean differences and pointwise 95\% confidence intervals are obtained from 20,000 paired dataset-level bootstrap resamples.}

{Table~\ref{tab:selectarena_significance} shows that PACE-DC and PACE-HSIC achieve consistent improvements across both evaluations. PACE-DC yields gains of 2.94 and 3.14 percentage points on the full benchmark and classification subset, respectively, while PACE-HSIC achieves gains of 1.71 and 1.75 percentage points. The improvement from PACE-SIS is relatively small. These results further support the effectiveness of pretrained representations in enhancing classical feature screening on natural datasets.}

\begin{table}[t]
\centering
\caption{\textbf{Paired PACE--Raw effects on SelectArena.} Effects and confidence intervals are in percentage points of error reduction from Random. Intervals are pointwise 95\% paired dataset-bootstrap intervals.}
\label{tab:selectarena_significance}
\small
\begin{tabular}{llrr}
\toprule
Evaluation & Utility & Mean effect & 95\% CI \\
\midrule
All tasks (71 datasets) & SIS & $-0.08$ & $[-2.23, +1.61]$ \\
 & DC & $+2.94$ & $[+1.55, +4.50]$ \\
 & HSIC & $+1.71$ & $[+0.52, +3.15]$ \\
\midrule
Classification (61 datasets) & SIS & $+0.54$ & $[-0.94, +2.01]$ \\
 & DC & $+3.14$ & $[+1.65, +4.77]$ \\
 & HSIC & $+1.75$ & $[+0.58, +3.11]$ \\
\bottomrule
\end{tabular}
\end{table}

%% file: appendix_real.tex
\section{Real-Data Evaluation Details and Additional Results}
\label{app:real_full}

\subsection{Study overview and evaluation protocol}
\label{app:real_protocol}

The real-data evaluation uses two complementary constructions.  The nuisance-contaminated TALENT study estimates matched raw-to-PACE representation effects across binary classification, multiclass classification, and regression.  The natural high-dimensional study evaluates end-to-end screening pipelines on tables with continuous features.

For TALENT, the \emph{clean table} is the original table, the \emph{nuisance-contaminated table} adds independent $\mathcal N(0,1)$ features until its total width is 2,000, and the \emph{screened table} contains the $k=\min(N-1,3p_{\mathrm{orig}})$ highest-ranked features, where $N$ is the training-set size. Raw-to-PACE contrasts match the selection budget, downstream learner, base dataset, and repeated split. Binary tasks use AUC, multiclass tasks use macro-averaged one-versus-rest AUC, and regression tasks use RMSE. Three random seeds are used. Representation extraction and utility estimation use at most 2,000 training rows: classification subsampling is stratified and regression subsampling is without replacement, both controlled by the corresponding random seed. 

The natural high-dimensional experiment uses ten datasets and five random seeds, retaining the top 30\% of features ranked using the training partition. Dataset-level results require all five seeds. For TabICL-SHAP and SAGE-CatBoost, an incomplete result indicates that at least one seed exceeded two hours; completed cells report mean feature-selection time directly in Table~\ref{tab:highdim_accuracy}. TabPFN-Wide and GOTabPFN operate directly on the full table and have complete coverage on the same nine-dataset support. On 20-class COIL20, all six PACE pipelines complete all five seeds; both dedicated wide predictors have incomplete five-seed coverage. Scores are averaged within each dataset, and each dataset receives equal weight in complete-coverage means.

\subsection{Nuisance-contaminated TALENT: matched representation effects}
\label{app:talent_matched}

Tables~\ref{tab:talent_learner_binary} and~\ref{tab:talent_learner_multiclass} report the classification effects by utility and downstream learner.

\begin{table}[!htbp]
\centering
\caption{\textbf{Learner-conditioned PACE-minus-raw effects: binary classification.} Each cell is the mean AUC gain followed by a 95\% base-dataset bootstrap interval. The intervention replaces a raw utility with its PACE counterpart at the same screening budget; positive values favor PACE.}
\label{tab:talent_learner_binary}
\small
\begin{tabular}{lrrr}
\toprule
Learner & SIS & DC & HSIC\\
\midrule
LightGBM & +0.010 [+0.004, +0.017] & +0.001 [-0.003, +0.003] & +0.008 [+0.003, +0.013]\\
XGBoost & +0.010 [+0.004, +0.017] & +0.003 [+0.001, +0.006] & +0.009 [+0.005, +0.014]\\
CatBoost & +0.011 [+0.004, +0.020] & +0.003 [+0.001, +0.006] & +0.004 [+0.001, +0.008]\\
MLP & +0.007 [+0.003, +0.011] & +0.002 [-0.000, +0.004] & +0.008 [+0.004, +0.012]\\
RealMLP & +0.012 [+0.006, +0.020] & +0.011 [+0.006, +0.017] & +0.012 [+0.005, +0.020]\\
TabM & +0.006 [+0.003, +0.010] & +0.002 [+0.000, +0.003] & +0.007 [+0.004, +0.011]\\
TabPFN v3 & +0.011 [+0.005, +0.020] & +0.002 [-0.001, +0.004] & +0.010 [+0.005, +0.016]\\
TabPFN v2.6 & +0.009 [+0.003, +0.017] & +0.002 [-0.002, +0.005] & +0.007 [-0.001, +0.014]\\
TabICL v2 & +0.012 [+0.005, +0.020] & +0.002 [-0.002, +0.006] & +0.014 [+0.007, +0.021]\\
TabDPT & +0.009 [+0.004, +0.014] & +0.004 [+0.002, +0.006] & +0.013 [+0.008, +0.019]\\
\bottomrule
\end{tabular}
\end{table}

\begin{table}[!htbp]
\centering
\caption{\textbf{Learner-conditioned PACE-minus-raw effects: multiclass classification.} Each cell is the mean macro-AUC gain followed by a 95\% base-dataset bootstrap interval. The intervention replaces a raw utility with its PACE counterpart at the same screening budget; positive values favor PACE.}
\label{tab:talent_learner_multiclass}
\small
\begin{tabular}{lrrr}
\toprule
Learner & SIS & DC & HSIC\\
\midrule
LightGBM & +0.008 [+0.003, +0.014] & -0.000 [-0.002, +0.002] & +0.004 [+0.001, +0.007]\\
XGBoost & +0.010 [+0.003, +0.020] & +0.004 [+0.003, +0.006] & +0.004 [+0.002, +0.006]\\
CatBoost & +0.009 [+0.002, +0.020] & +0.000 [-0.001, +0.001] & +0.003 [+0.001, +0.005]\\
MLP & +0.010 [+0.003, +0.019] & +0.000 [-0.002, +0.002] & +0.010 [+0.005, +0.014]\\
RealMLP & +0.010 [+0.002, +0.020] & +0.001 [-0.003, +0.005] & +0.004 [-0.0001, +0.009]\\
TabM & +0.012 [+0.004, +0.022] & +0.002 [-0.003, +0.007] & +0.002 [-0.001, +0.004]\\
TabPFN v3 & +0.013 [+0.004, +0.025] & +0.000 [-0.001, +0.001] & +0.004 [+0.002, +0.006]\\
TabPFN v2.6 & +0.012 [+0.004, +0.025] & +0.000 [-0.001, +0.001] & +0.004 [+0.002, +0.007]\\
TabICL v2 & +0.013 [+0.004, +0.025] & +0.003 [+0.002, +0.004] & +0.005 [+0.002, +0.008]\\
TabDPT & +0.008 [+0.003, +0.017] & +0.002 [+0.001, +0.003] & +0.005 [+0.002, +0.007]\\
\bottomrule
\end{tabular}
\end{table}

\begin{table}[!htbp]
\centering
\caption{\textbf{Nuisance damage and PACE gains by model family.} Damage is clean-minus-contaminated binary/multiclass AUC for classification and the clean-normalized RMSE increase for regression. PACE gains average matched PACE-minus-raw effects equally over SIS, DC, and HSIC and the learners within each family, using dataset-mean scores. We report means across datasets for classification and medians for regression, with 95\% dataset-bootstrap intervals. Positive values indicate greater damage or gain.}
\label{tab:talent_family_sensitivity}
\scriptsize
\resizebox{\linewidth}{!}{%
\begin{tabular}{lrrrrrr}
\toprule
& \multicolumn{2}{c}{Binary} & \multicolumn{2}{c}{Multiclass} & \multicolumn{2}{c}{Regression}\\
\cmidrule(lr){2-3}\cmidrule(lr){4-5}\cmidrule(lr){6-7}
Family & Nuisance damage & PACE effect & Nuisance damage & PACE effect & Nuisance damage & PACE effect\\
\midrule
Tree & $+0.049\;[+0.039,+0.059]$ & $+0.0069\;[+0.0043,+0.0100]$ & $+0.028\;[+0.020,+0.037]$ & $+0.0034\;[+0.0016,+0.0059]$ & $+0.118\;[+0.076,+0.188]$ & $+0.0004\;[-0.0026,+0.0034]$\\
NN & $+0.124\;[+0.112,+0.137]$ & $+0.0075\;[+0.0053,+0.0100]$ & $+0.119\;[+0.101,+0.138]$ & $+0.0042\;[+0.0022,+0.0064]$ & $+0.419\;[+0.281,+0.599]$ & $+0.0008\;[-0.0011,+0.0035]$\\
TFM & $+0.114\;[+0.099,+0.131]$ & $+0.0080\;[+0.0050,+0.0115]$ & $+0.081\;[+0.067,+0.096]$ & $+0.0042\;[+0.0023,+0.0067]$ & $+0.633\;[+0.366,+1.879]$ & $+0.0009\;[-0.0027,+0.0041]$\\
\bottomrule
\end{tabular}}
\end{table}

\paragraph{Sensitivity across downstream model families.}
Table~\ref{tab:talent_family_sensitivity} reports nuisance damage and PACE gains averaged over SIS, DC, and HSIC for trees, NNs, and TFMs. 

\paragraph{Contamination recovery.}
Let $Q$ denote binary AUC or multiclass macro-AUC and $R$ denote RMSE.  For each matched learner--dataset--split combination, the restored fraction of nuisance damage is
\[
\frac{Q_{\mathrm{screened}}-Q_{\mathrm{contaminated}}}{Q_{\mathrm{clean}}-Q_{\mathrm{contaminated}}}
\quad\text{or}\quad
\frac{R_{\mathrm{contaminated}}-R_{\mathrm{screened}}}{R_{\mathrm{contaminated}}-R_{\mathrm{clean}}},
\]
where the subscripts denote the clean, nuisance-contaminated, and screened tables.  The estimand is the matched PACE-minus-raw recovery difference for the same utility.  Ratios are evaluated when nuisance damage exceeds $\epsilon$: damage is $Q_{\mathrm{clean}}-Q_{\mathrm{contaminated}}$ for classification and $(R_{\mathrm{contaminated}}-R_{\mathrm{clean}})/R_{\mathrm{clean}}$ for regression. Recovery differences are averaged over learners within each base dataset before bootstrap resampling.

\begin{figure}[!htbp]
\centering
\includegraphics[width=.99\linewidth]{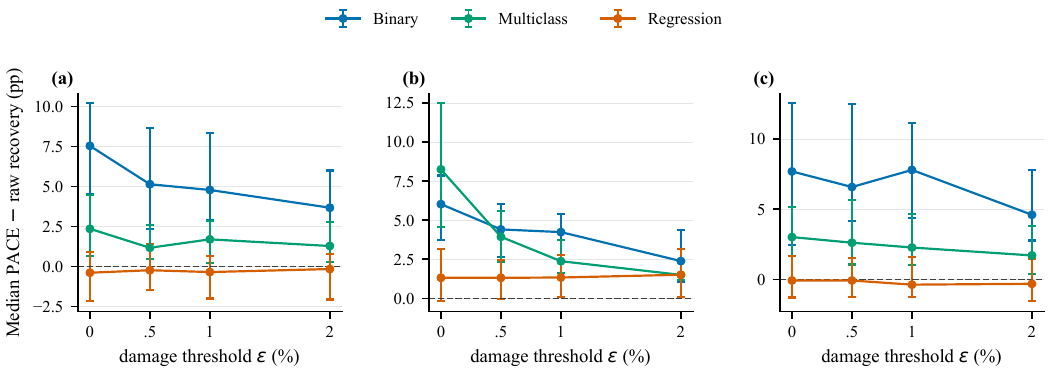}
\caption{\textbf{Contamination recovery across damage thresholds.} Panels (a)--(c) show SIS, DC, and HSIC\@. Points are median PACE-minus-raw recovery differences with 95\% bootstrap intervals; thresholds are absolute classification-score loss or regression loss relative to clean-table RMSE\@. Intervals lie above zero for all classification utilities; regression DC intervals include zero at the two lowest thresholds, while regression SIS and HSIC remain near zero.}
\label{fig:contamination_recovery}
\end{figure}

\begin{table}[!htbp]
\centering
\caption{\textbf{Contamination recovery at $\epsilon=.01$.} Recovery is the fraction of performance loss caused by nuisance contamination that is restored, computed per matched split--learner--dataset combination before aggregation. Raw/PACE columns report marginal medians. Column 5 reports the median paired recovery difference in percentage points with a 95\% bootstrap interval; the final column gives its native-scale counterpart. Ratios are unclipped.}
\label{tab:contamination_recovery}
\small
\begin{tabular}{llrrrr}
\toprule
Task & Utility & Raw recovery & PACE recovery & PACE $-$ raw [95\% interval] & Native-scale $\Delta$\\
\midrule
Binary & SIS & 0.382 & 0.528 & +4.79 [+2.92, +8.37] & +0.0027\\
Binary & DC & 0.509 & 0.610 & +4.24 [+2.31, +5.41] & +0.0014\\
Binary & HSIC & 0.497 & 0.652 & +7.80 [+4.37, +11.17] & +0.0035\\
\midrule
Multiclass & SIS & 0.723 & 0.739 & +1.70 [+0.24, +2.83] & +0.0009\\
Multiclass & DC & 0.741 & 0.788 & +2.38 [+1.61, +3.71] & +0.0011\\
Multiclass & HSIC & 0.722 & 0.777 & +2.27 [+1.03, +4.64] & +0.0020\\
\midrule
Regression & SIS & 0.569 & 0.573 & -0.35 [-2.00, +0.65] & 0.0000\\
Regression & DC & 0.751 & 0.784 & +1.33 [+0.07, +2.76] & +0.0023\\
Regression & HSIC & 0.638 & 0.649 & -0.38 [-1.25, +1.59] & 0.0000\\
\bottomrule
\end{tabular}
\end{table}

Figure~\ref{fig:contamination_recovery} and Table~\ref{tab:contamination_recovery} report recovery differences across damage thresholds. 

\FloatBarrier

\begingroup

\subsection{Row-permuted nuisance features}
\label{app:talent_row_permutation}

We repeat the TALENT classification experiment using independently row-permuted copies of original features as nuisance columns. Each table is extended to 2,000 features, with permutations performed separately within the training, validation, and test splits. This preserves each source feature's empirical distribution while randomizing its alignment with the response. We evaluate 98 binary and 57 multiclass datasets with fewer than 20,000 observations and at most ten classes, using three seeds and LightGBM, MLP, RealMLP, and TabDPT as downstream predictors. Raw and PACE use the same budget, $k=\min(3p_{\mathrm{orig}},N-1,2000)$.

\begin{table}[!htbp]
\centering

\caption{\textbf{Screening under row-permuted nuisance.} Retention is the fraction of original features retained. Predictive effects are PACE-minus-raw AUC gains (macro-AUC for multiclass), averaged over seeds and learners within each dataset and then equally across datasets. Brackets give 95\% intervals from 10,000 dataset-bootstrap resamples.}
\label{tab:talent_row_permutation}
\small
\begin{tabular}{llcc}
\toprule
Task & Utility & Retention: raw $\to$ PACE & AUC gain [95\% CI]\\
\midrule
Binary & SIS & $0.6127\to0.6981$ & $+0.0111\;[0.0053,0.0182]$\\
 & DC & $0.6834\to0.7094$ & $+0.0041\;[0.0009,0.0080]$\\
 & HSIC & $0.6607\to0.6741$ & $+0.0068\;[0.0031,0.0107]$\\
\midrule
Multiclass & SIS & $0.7860\to0.8290$ & $+0.0033\;[0.0000,0.0066]$\\
 & DC & $0.8108\to0.8189$ & $+0.0020\;[-0.0004,0.0048]$\\
 & HSIC & $0.7934\to0.7970$ & $+0.0042\;[0.0017,0.0067]$\\
\bottomrule
\end{tabular}
\end{table}

Table~\ref{tab:talent_row_permutation} shows higher original-feature retention for all three utilities in both tasks. SIS yields the largest retention gains, at 8.54 and 4.30 percentage points for binary and multiclass classification. Predictive gains remain positive on average, with intervals above zero for all three binary comparisons and multiclass HSIC. PACE thus remains useful when nuisance columns share the marginal distributions of original features.
\endgroup
\FloatBarrier

\subsection{TALENT screening comparison: rank, retention, and time}
\label{app:talent_field}

Main-text Figure~\ref{fig:real_rank} presents the performance--time comparisons and the common-support pairwise matrix. Table~\ref{tab:talent_full_field} provides task-specific ranks for the 18 ranked methods and records the five methods exceeding the three-hour screening budget.

\begin{table}[!htbp]
\centering
\caption{\textbf{TALENT comparison across 23 screening methods.} Ranks are computed within learner--dataset blocks and averaged by dataset; lower is better. Methods exceeding the three-hour screening budget are marked $>3\,\mathrm{h}$ and excluded from ranking. Retention is the task-equal mean over available panels. Runtimes are in seconds unless labeled in hours; dashes denote unavailable or inapplicable results.}
\label{tab:talent_full_field}
\scriptsize
\resizebox{\linewidth}{!}{%
\begin{tabular}{llrrrrr}
\toprule
Method & Family & Binary rank & Multiclass rank & Regression rank & Retention (\%) & Time\\
\midrule
Raw SIS & Raw utility & 10.62 & 9.62 & 9.38 & 66.4 & 0.14\\
\textbf{PACE-SIS} & PACE & 7.73 & 7.74 & 9.99 & 71.5 & 0.21\\
Raw DC & Raw utility & 9.08 & 9.49 & 6.90 & 73.5 & 0.98\\
\textbf{PACE-DC} & PACE & 7.36 & 6.46 & 5.62 & 75.1 & 1.12\\
Raw HSIC & Raw utility & 9.03 & 8.54 & 7.98 & 68.7 & 11.39\\
\textbf{PACE-HSIC} & PACE & 5.94 & 6.51 & 7.37 & 70.7 & 6.60\\
\midrule
IDS & Fixed comparator & 12.85 & 11.12 & 12.88 & 52.5 & 0.19\\
\midrule
RF-MDI & Model-dependent & 10.12 & 9.46 & 8.81 & 55.5 & 52.89\\
LGB-Split & Model-dependent & 9.10 & 9.22 & 8.09 & 57.5 & 8.41\\
LGB-Gain & Model-dependent & 8.85 & 9.40 & 8.26 & 57.0 & 8.12\\
CatBoost-PVC & Model-dependent & 8.95 & 9.39 & 8.20 & 55.4 & 110.85\\
\midrule
LGB-SHAP & Predictive attribution & 10.51 & 9.60 & 8.48 & 56.3 & 30.97\\
CatBoost-SHAP & Predictive attribution & 9.45 & 9.64 & 8.44 & 57.0 & 119.55\\
TabICL-SHAP & Predictive attribution & -- & -- & -- & -- & $>3\,\mathrm{h}$\\
LOCO-simple & Predictive attribution & -- & -- & -- & -- & $>3\,\mathrm{h}$\\
LOCO-CV & Predictive attribution & -- & -- & -- & -- & $>3\,\mathrm{h}$\\
SAGE-LGB & Predictive attribution & -- & -- & -- & -- & $>3\,\mathrm{h}$\\
SAGE-CatBoost & Predictive attribution & -- & -- & -- & -- & $>3\,\mathrm{h}$\\
\midrule
xRFM-AGOP & Learned selector & 9.54 & 9.94 & 9.17 & 66.3 & 1.69\\
LassoNet & Learned selector & 11.39 & 10.71 & 13.14 & 51.9 & 46.75\\
GradEnFS & Learned selector & 11.04 & 12.76 & -- & 48.4 & 4.42\\
SAND & Learned selector & 9.59 & 10.70 & 10.00 & 59.1 & 6.08\\
Deep Lasso & Learned selector & 9.86 & 10.69 & 10.29 & 60.0 & 7.33\\
\bottomrule
\end{tabular}}
\end{table}

Within each downstream learner and base dataset, methods are ranked after repeated-split aggregation; learner-specific ranks are then averaged within each dataset. Improvability and pairwise win rates are computed by averaging over learners within datasets and then weighting tasks equally; GradEnFS is evaluated only on classification. Retention summaries use the available task-specific panels, and timing summaries use each method's available TALENT base datasets.

{In Table~\ref{tab:talent_full_field}, PACE-HSIC achieves the best average rank for binary classification ($5.94$), while PACE-DC achieves the best average ranks for multiclass classification ($6.46$) and regression ($5.62$). PACE-HSIC obtains average ranks of $6.51$ and $7.37$ for multiclass classification and regression, respectively. Notably, raw DC is already a strong baseline with high retention ($73.5\%$), especially for regression, where its rank of $6.90$ exceeds most methods; PACE-DC still improves this rank to $5.62$ and raises retention to $75.1\%$.}

\paragraph{Feature-dimension timing.}
\label{app:runtime_scaling_protocol}
For Figure~\ref{fig:runtime_tfm_comparison}(c), GPU-based computations use an NVIDIA GeForce RTX 4080 SUPER. LOCO-simple is run with 650 training and 350 validation observations from the same $N=1{,}000$ synthetic-data construction. Each run fits one full random forest and $p$ reduced forests, scoring every feature by its held-out log-loss increase after deletion. Forests have 300 trees, a maximum depth of six, and a candidate-feature fraction of $0.8$. The curve averages two repetitions at each completed width. Timing includes model fitting, prediction, and loss evaluation, and excludes data generation and checkpoint I/O. Triangles indicate completed measurements above the three-hour display ceiling.

\FloatBarrier

\subsection{Runtime scaling with sample size}
\label{app:sample_runtime_scaling}

We fix $p=1{,}000$ and vary $N\in\{100,200,500,1{,}000,2{,}000,5{,}000\}$ under the same data-generation and timing protocol as the feature-dimension experiment. 

\begin{figure}[!htbp]
\centering
\includegraphics[width=.9\linewidth]{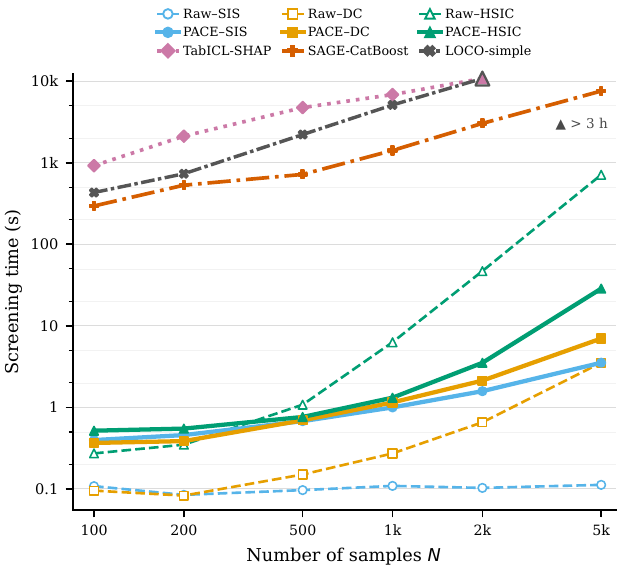}
\caption{\textbf{Screening time versus sample size at $p=1{,}000$.} Solid and dashed curves denote PACE and raw utilities. Nested triangles mark the $>3$-hour budget bounds for TabICL-SHAP and LOCO-simple at $N=2{,}000$.}
\label{fig:sample_runtime_scaling}
\end{figure}

PACE remains substantially faster than the attribution methods across the measured sample sizes (Figure~\ref{fig:sample_runtime_scaling}). At $N=5{,}000$, PACE-SIS, PACE-DC, and PACE-HSIC take $3.53$, $6.99$, and $28.61$ seconds, respectively, compared with $7{,}582$ seconds for SAGE-CatBoost---speedups of approximately $2{,}149\times$, $1{,}084\times$, and $265\times$. TabICL-SHAP and LOCO-simple exceed the three-hour budget at $N=2{,}000$.

PACE-SIS grows gradually with $N$, while DC and HSIC become more expensive as pairwise computations increase. DC utility evaluation grows approximately quadratically at larger sample sizes; HSIC grows more steeply because its current implementation includes dense kernel-matrix products. SAGE-CatBoost grows roughly linearly over the larger settings but remains much slower. Distributed computation, as explored in CRU and LR-FFS~\citep{li2024feature,qin2025label}, offers a further route to accelerating screening on large samples.

\FloatBarrier

\subsection{Natural high-dimensional classification}
\label{app:highdim_classification}

The natural high-dimensional panel contains five binary and five multiclass datasets with $500$--$10{,}304$ features.  Screened pipelines retain 30\% of the features ranked on the training partition and are evaluated on the same five splits using TabPFN v3 and TabICL v2.  Table~\ref{tab:highdim_accuracy} compares full-table prediction with matched raw and PACE versions of SIS, DC, and HSIC; IDS; learned and attribution-based selectors; LightGBM gain importance; and the end-to-end TabPFN-Wide and GOTabPFN predictors.

\begin{table}[!htbp]
\centering
\caption{\textbf{Natural high-dimensional classification on ten datasets.} Screeners retain the top 30\% of training-ranked features; Wide/Go are end-to-end TabPFN-Wide/GOTabPFN. Entries are five-split mean accuracy (\%), ordered by $p/n$. Means weight all ten datasets equally and require complete coverage; em dashes mark incomplete coverage, and parenthetical T-SHAP/SAGE values give selection time (missing if any split exceeds two hours). All PACE pipelines complete 20-class COIL20. Bold marks the best result for each dataset--backbone pair and each mean row. LNet, Grad, D-L, T-SHAP, and SAGE denote LassoNet, GradEnFS, Deep Lasso, TabICL-SHAP, and SAGE-CatBoost.}
\label{tab:highdim_accuracy}
\setlength{\tabcolsep}{1.5pt}
\renewcommand{\arraystretch}{1.03}
\scriptsize
\resizebox{\textwidth}{!}{%
\begin{tabular}{@{}l r c cc cc cc *{11}{c}@{}}
\toprule
Dataset & $p/n$ & Full & \multicolumn{2}{c}{SIS} & \multicolumn{2}{c}{DC} & \multicolumn{2}{c}{HSIC} & IDS & xRFM-AGOP & LNet & Grad & SAND & D-L & T-SHAP & SAGE & LGB-Gain & Wide & Go \\
\cmidrule(lr){4-5}\cmidrule(lr){6-7}\cmidrule(lr){8-9}
 & & & Raw & PACE & Raw & PACE & Raw & PACE & & & & & & & & & & & \\
\midrule
\multicolumn{20}{@{}l}{\textbf{TabPFN v3}} \\
orlraws10P & 103.0 & 99.0 & 97.0 & 99.0 & \textbf{100.0} & 99.0 & 98.0 & \textbf{100.0} & 97.0 & 99.0 & 99.0 & 98.0 & 98.0 & 98.0 & \textemdash~($>2\,\mathrm{h}$) & \textemdash~($>2\,\mathrm{h}$) & 98.0 & \textbf{100.0} & 99.0 \\
leukemia & 98.2 & 92.0 & 93.3 & 93.3 & 93.3 & 93.3 & 93.3 & 93.3 & 92.0 & 93.3 & 92.0 & 93.3 & 93.3 & 92.0 & 93.3~($40.7\,\mathrm{min}$) & \textemdash~($>2\,\mathrm{h}$) & 93.3 & \textbf{94.7} & 86.7 \\
GLIOMA & 88.7 & 64.0 & 64.0 & \textbf{70.0} & 64.0 & 64.0 & 60.0 & 60.0 & 62.0 & 60.0 & 66.0 & 60.0 & 66.0 & 64.0 & 62.0~($9.8\,\mathrm{min}$) & \textemdash~($>2\,\mathrm{h}$) & 66.0 & 64.0 & 60.0 \\
arcene & 50.0 & 88.5 & 85.5 & 89.0 & 88.0 & 89.5 & 90.0 & 86.5 & 87.5 & 83.0 & 88.0 & \textbf{92.0} & 86.0 & 88.0 & \textemdash~($>2\,\mathrm{h}$) & \textemdash~($>2\,\mathrm{h}$) & 90.0 & 89.5 & 89.0 \\
lung & 16.3 & 93.7 & 94.6 & 95.6 & 95.6 & 94.1 & 95.6 & 95.1 & 95.6 & 94.1 & 95.6 & 93.7 & \textbf{96.1} & 95.1 & \textbf{96.1}~($16.7\,\mathrm{min}$) & \textemdash~($>2\,\mathrm{h}$) & 95.1 & 94.6 & 93.7 \\
warpPIE10P & 11.5 & \textbf{100.0} & \textbf{100.0} & \textbf{100.0} & \textbf{100.0} & \textbf{100.0} & \textbf{100.0} & \textbf{100.0} & \textbf{100.0} & \textbf{100.0} & \textbf{100.0} & \textbf{100.0} & \textbf{100.0} & \textbf{100.0} & \textbf{100.0}~($10.7\,\mathrm{min}$) & \textbf{100.0}~($1.44\,\mathrm{h}$) & \textbf{100.0} & \textbf{100.0} & \textbf{100.0} \\
BASEHOCK & 2.4 & 89.3 & 95.5 & 96.5 & 96.7 & 96.0 & 96.5 & 96.4 & 96.0 & 95.0 & \textemdash & 95.9 & 95.9 & 94.8 & \textemdash~($>2\,\mathrm{h}$) & \textemdash~($>2\,\mathrm{h}$) & 93.8 & \textbf{98.3} & 97.2 \\
PCMAC & 1.7 & 83.1 & 90.5 & 89.5 & 90.5 & 91.1 & 89.8 & 90.2 & 89.9 & 89.7 & \textemdash & 88.7 & 89.8 & 91.2 & 90.6~($1.85\,\mathrm{h}$) & \textemdash~($>2\,\mathrm{h}$) & 89.0 & \textbf{92.0} & 89.0 \\
COIL20 & 0.7 & \textbf{100.0} & \textbf{100.0} & \textbf{100.0} & \textbf{100.0} & \textbf{100.0} & \textbf{100.0} & 99.9 & \textbf{100.0} & \textbf{100.0} & \textbf{100.0} & \textbf{100.0} & \textbf{100.0} & \textbf{100.0} & \textbf{100.0}~($24.0\,\mathrm{min}$) & \textbf{100.0}~($1.86\,\mathrm{h}$) & \textbf{100.0} & \textemdash & \textemdash \\
madelon & 0.2 & 89.5 & 89.9 & 90.1 & 90.1 & 89.9 & 90.1 & 90.1 & 70.8 & \textbf{90.4} & 77.4 & 89.5 & 90.1 & 87.8 & 90.2~($10.7\,\mathrm{min}$) & 90.3~($33.6\,\mathrm{min}$) & 90.1 & 87.4 & 67.7 \\
\cmidrule(lr){1-20}
Mean & \textemdash & 89.9 & 91.0 & \textbf{92.3} & 91.8 & 91.7 & 91.3 & 91.2 & 89.1 & 90.5 & \textemdash & 91.1 & 91.5 & 91.1 & \textemdash & \textemdash & 91.5 & \textemdash & \textemdash \\
\midrule
\multicolumn{20}{@{}l}{\textbf{TabICL v2}} \\
orlraws10P & 103.0 & \textbf{100.0} & 98.0 & \textbf{100.0} & 98.0 & \textbf{100.0} & 99.0 & \textbf{100.0} & 98.0 & 98.0 & 99.0 & 99.0 & \textbf{100.0} & 98.0 & \textemdash~($>2\,\mathrm{h}$) & \textemdash~($>2\,\mathrm{h}$) & 99.0 & \textbf{100.0} & 99.0 \\
leukemia & 98.2 & 90.7 & \textbf{97.3} & 96.0 & 94.7 & 94.7 & \textbf{97.3} & 94.7 & \textbf{97.3} & 94.7 & 94.7 & 94.7 & 96.0 & 96.0 & 96.0~($40.7\,\mathrm{min}$) & \textemdash~($>2\,\mathrm{h}$) & 90.7 & 94.7 & 86.7 \\
GLIOMA & 88.7 & 52.0 & 58.0 & 64.0 & \textbf{66.0} & \textbf{66.0} & 62.0 & 62.0 & 62.0 & 60.0 & 60.0 & 62.0 & 64.0 & 54.0 & 56.0~($9.8\,\mathrm{min}$) & \textemdash~($>2\,\mathrm{h}$) & 56.0 & 64.0 & 60.0 \\
arcene & 50.0 & 82.0 & 88.0 & 86.5 & 88.5 & 87.0 & 87.0 & 87.5 & 84.5 & 75.0 & 75.5 & 85.5 & 83.0 & 80.0 & \textemdash~($>2\,\mathrm{h}$) & \textemdash~($>2\,\mathrm{h}$) & 84.0 & \textbf{89.5} & 89.0 \\
lung & 16.3 & 94.6 & 95.1 & 94.1 & 95.1 & \textbf{95.6} & \textbf{95.6} & 95.1 & 94.1 & 94.6 & \textbf{95.6} & 94.6 & \textbf{95.6} & \textbf{95.6} & 95.1~($16.7\,\mathrm{min}$) & \textemdash~($>2\,\mathrm{h}$) & 94.6 & 94.6 & 93.7 \\
warpPIE10P & 11.5 & \textbf{100.0} & \textbf{100.0} & \textbf{100.0} & \textbf{100.0} & \textbf{100.0} & \textbf{100.0} & \textbf{100.0} & \textbf{100.0} & \textbf{100.0} & \textbf{100.0} & \textbf{100.0} & \textbf{100.0} & \textbf{100.0} & \textbf{100.0}~($10.7\,\mathrm{min}$) & \textbf{100.0}~($1.44\,\mathrm{h}$) & \textbf{100.0} & \textbf{100.0} & \textbf{100.0} \\
BASEHOCK & 2.4 & 93.1 & 98.2 & 98.0 & 98.2 & 98.4 & 98.2 & 98.2 & 98.5 & 98.2 & \textemdash & 97.9 & 98.3 & \textbf{98.6} & \textemdash~($>2\,\mathrm{h}$) & \textemdash~($>2\,\mathrm{h}$) & 97.6 & 98.3 & 97.2 \\
PCMAC & 1.7 & 90.7 & 94.2 & \textbf{94.3} & 93.9 & 93.7 & 93.8 & 93.8 & 94.0 & 93.6 & \textemdash & 92.5 & 94.2 & 93.9 & 93.3~($1.85\,\mathrm{h}$) & \textemdash~($>2\,\mathrm{h}$) & 91.6 & 92.0 & 89.0 \\
COIL20 & 0.7 & \textbf{100.0} & \textbf{100.0} & \textbf{100.0} & \textbf{100.0} & \textbf{100.0} & \textbf{100.0} & \textbf{100.0} & \textbf{100.0} & \textbf{100.0} & \textbf{100.0} & \textbf{100.0} & \textbf{100.0} & \textbf{100.0} & \textbf{100.0}~($24.0\,\mathrm{min}$) & \textbf{100.0}~($1.86\,\mathrm{h}$) & \textbf{100.0} & \textemdash & \textemdash \\
madelon & 0.2 & 89.5 & \textbf{89.9} & 89.5 & 89.4 & 89.7 & \textbf{89.9} & 89.2 & 74.8 & 89.7 & 77.0 & 89.4 & 89.3 & 88.5 & 89.8~($10.7\,\mathrm{min}$) & 89.6~($33.6\,\mathrm{min}$) & 89.5 & 87.4 & 67.7 \\
\cmidrule(lr){1-20}
Mean & \textemdash & 89.3 & 91.9 & 92.3 & 92.4 & \textbf{92.5} & 92.3 & 92.1 & 90.3 & 90.4 & \textemdash & 91.6 & 92.0 & 90.5 & \textemdash & \textemdash & 90.3 & \textemdash & \textemdash \\
\bottomrule
\end{tabular}%
}
\end{table}

On the ten evaluated datasets, Table~\ref{tab:highdim_accuracy} shows that a PACE utility has the highest complete-coverage mean with each downstream TFM, while effects vary by utility and dataset. Under TabPFN v3, PACE-SIS reaches $0.923$, compared with $0.910$ for raw SIS and $0.899$ for the full table, and LightGBM gain and SAND share the highest complete-coverage accuracy among learned or model-based selectors, at $0.915$. Under TabICL v2, PACE-DC reaches $0.925$, compared with $0.924$ for raw DC, $0.923$ for raw HSIC, $0.920$ for SAND, and $0.893$ for the full table.

All six raw/PACE utility pipelines in Table~\ref{tab:highdim_accuracy} complete all ten datasets for both downstream TFMs. On the nine datasets shared by the dedicated wide predictors, PACE-DC with TabICL v2 reaches $0.917$, compared with $0.912$ for TabPFN-Wide and $0.869$ for GOTabPFN; PACE-SIS with TabPFN v3 reaches $0.914$. On 20-class COIL20, the six PACE pipelines complete all five seeds, with mean accuracies ranging from $99.93\%$ to $100\%$, whereas both dedicated wide predictors are incomplete.

For the same panel in Table~\ref{tab:highdim_accuracy}, PACE-SIS completes feature screening in $30.36$ seconds on average across datasets and splits, compared with the two-hour per-split execution cap for TabICL-SHAP and SAGE-CatBoost.

{We additionally tune GOTabPFN on the nine datasets with complete coverage, using five splits per dataset. Optuna's TPE sampler searches the authors' 14-parameter binary-classification search space for both classification tasks, with up to 50 trials and 360 seconds per split. Hyperparameters are selected by validation accuracy; the model is then refitted on the training and validation partitions and evaluated on the held-out test partition, using all original features. Tuning raises mean accuracy from $86.94\%$ to $88.97\%$. On the same datasets, PACE-SIS retaining 30\% of features achieves $91.44\%$ with TabPFN v3 and $91.39\%$ with TabICL v2. The paired mean advantages over tuned GOTabPFN are $2.48$ percentage points (95\% CI: $[0.14,5.32]$) and $2.43$ points ($[0.32,5.20]$), respectively, averaging splits within datasets and bootstrapping datasets.}

\FloatBarrier